\documentclass[journal]{IEEEtran}

\usepackage{amsmath,amssymb}
\usepackage{colortbl}
\usepackage{diagbox}
\usepackage{subeqnarray}
\usepackage{cases}
\usepackage{setspace}
\usepackage{tabularx}
\usepackage{booktabs}
\usepackage{multicol}
\usepackage{dcolumn}
\usepackage{graphicx}
\usepackage{subfigure}
\usepackage{float}
\usepackage{epsfig}
\usepackage{amsmath}
\usepackage{amssymb}
\usepackage{array}
\usepackage{epstopdf}
\usepackage{color}
\usepackage{xcolor}
\usepackage{psfrag}
\usepackage{cite}
\usepackage{multirow}
\usepackage{color}
\usepackage{bm}
\usepackage{adjustbox}
\usepackage{overpic}
\usepackage{rotating}
\usepackage{marvosym}

\graphicspath{{figures/}}
\usepackage[bookmarks,colorlinks,linkcolor=red, anchorcolor=blue, citecolor=green]{hyperref}
\begin{document}
	
	\title{Learning Class-Agnostic Pseudo Mask Generation for Box-Supervised Semantic Segmentation}
	
	%
	
	\author{Chaohao~Xie,
		Dongwei~Ren, \emph{Member, IEEE,}
		Lei~Wang,
		Wangmeng~Zuo, \emph{Senior Member, IEEE}
		
		\thanks{C. Xie, D. Ren, L. Wang and W. Zuo are with Harbin Institute of Technology, Harbin, 150001, China (e-mail: viousxie@outlook.com; rendongweihit@gmail.com; wangleihitcs@qq.com; wmzuo@hit.edu.cn).}
		
	}
	
	\maketitle
	
	\begin{abstract}
		Recently, several weakly supervised learning methods have been devoted to utilize bounding box supervision for training deep semantic segmentation models.
		Most existing methods usually leverage the generic proposal generators (\emph{e.g.}, dense CRF and MCG) to produce enhanced segmentation masks for further training segmentation models.
		These proposal generators, however, are generic and not specifically designed for box-supervised semantic segmentation, thereby leaving some leeway for improving segmentation performance.
		In this paper, we aim at seeking for a more accurate learning-based class-agnostic pseudo mask generator tailored to box-supervised semantic segmentation.
		To this end, we resort to an auxiliary dataset with pixel-level annotation while the class labels are non-overlapped with those of the box-annotated dataset.
		For learning pseudo mask generator from the auxiliary dataset, we present a bi-level optimization formulation.
		In particular, the lower subproblem is used to learn box-supervised semantic segmentation, while the upper subproblem is used to learn an optimal class-agnostic pseudo mask generator.
		The learned pseudo segmentation mask generator can then be deployed to the box-annotated dataset for improving weakly supervised semantic segmentation.
		Experiments on PASCAL VOC 2012 dataset show that the learned pseudo mask generator is effective in boosting segmentation performance, and our method can largely close the performance gap between box-supervised and fully-supervised models.

	\end{abstract}
	

	\IEEEpeerreviewmaketitle

	\section{Introduction}

	\IEEEPARstart{I}{mage} semantic segmentation is to label each pixel with specific semantic category.
	Benefited from the advances in deep learning, convolutional neural network (CNN)-based methods~\cite{kang2018depth,Lin_2016_CVPR,CRF_as_RNN,RNN_SceneLabel,chen_deeplabv2,zhao2017pspnet} have achieved unprecedented success on image segmentation.
	Fully Convolutional Networks (FCN)~\cite{Long_FCN_2015} and U-Net~\cite{UNetRFB15a} are two representative architectures for semantic segmentation.
	Hierarchical convolutional features~\cite{Ross_RCNN, HariharanAGM15}, feature pyramids ~\cite{Chen_Rethink,FPN_Lin_2017,zhao2017pspnet,xiao2018unified}, multi-path structure~\cite{fu2019stacked,ding2020semantic,geng2021gated}, and dilated convolution~\cite{chen_deeplabv2} also have shown to be beneficial to enhance feature representation ability.
	Nonetheless, state-of-the-art deep semantic segmentation models generally are trained in the fully supervised manner, and heavily depend on the laborious and costly pixel-level annotations of enormous images~\cite{pascal_voc,MSCOCO}.

	\begin{figure*}[ht]
		\small
		\centering
		\includegraphics[width=0.98\linewidth]{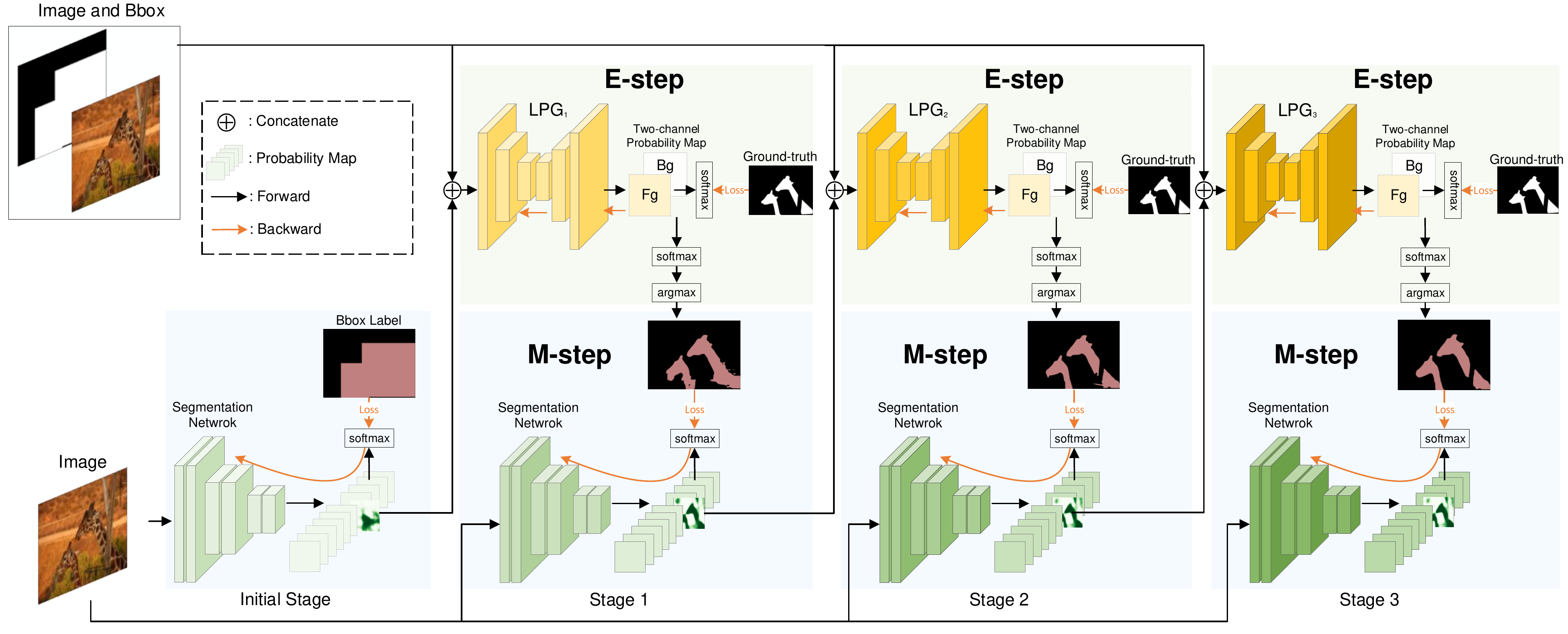}
		\caption{Illustration of learning stage-wise LPG models on auxiliary dataset $\mathcal{D}_{aux}$ by using the EM algorithm to solve bi-level optimization problem in Eqn. \eqref{upper}. 
			In the E-step, pixel-level annotations are adopted as ground-truth to train LPG. 
			In the M-step, a deep segmentation network is trained by using the pixel-level pseudo masks generated by LPG for supervision.  
			The alternating executions of the E-step and M-step result in stage-wise LPG, which can then be deployed to any box-annotated dataset $\mathcal{D}$ for generating accurate pixel-level segmentation mask and improving segmentation performance.   
		}
		\label{fig:figure2}
		\vspace{-1.0em}
	\end{figure*}

	Weak annotations, \emph{e.g.}, image-level labels~\cite{Ahn_2018_AffinityNet,Ahn_2019_IRNet,adver_erasing,hou2018selferasing}, points~\cite{whats_the_point,mcever2020pcams,extrem_cut}, scribbles~\cite{ScribbleSup,Regularized_loss} and bounding boxes~\cite{BoxSup_Dai_2015,Papandreou_2015_ICCV,Rajchl2017DeepCutOS,Khoreva_2017_SDI,hu2018learning,BCM_Song_2019_CVPR}, are much easier and cheaper to collect, but are less precise than pixel-level annotations.
	Weakly supervised semantic segmentation has then been investigated to close the gap to the fully-supervised counterpart~\cite{Papandreou_2015_ICCV,Ahn_2019_IRNet,mcever2020pcams,BCM_Song_2019_CVPR}.
	Among the weak annotations, bounding box annotation is very economical and can be readily obtained from object detection training sets, while still providing certain object localization information.
	Thus, several recent methods have been proposed to utilize bounding box annotation for weakly supervised semantic segmentation~\cite{BoxSup_Dai_2015,Papandreou_2015_ICCV,Khoreva_2017_SDI,BCM_Song_2019_CVPR}.

	Most existing methods usually leverage the generic proposal generators~\cite{DenseCRF,Roth_GrabCut,MCG_proposal} to produce enhanced
	segmentation proposals for further training segmentation models.
	For example, dense CRF~\cite{DenseCRF} is adopted in WSSL~\cite{Papandreou_2015_ICCV} and BCM~\cite{BCM_Song_2019_CVPR}, while GrabCut~\cite{Roth_GrabCut} and MCG~\cite{MCG_proposal} are combined in SDI \cite{Khoreva_2017_SDI} for generating more precise segmentation proposals. 
	Obviously, these generic proposal generators~\cite{DenseCRF,Roth_GrabCut,MCG_proposal} are not specifically designed for box-supervised semantic segmentation,
	while improved segmentation performance can be expected by using more precise proposal generator.
	Moreover, the gap between box and fully supervised learning can be completely eliminated when the proposal generator can produce the ground-truth segmentation masks.
	Thus, it is appealing to find a better segmentation mask generator tailored to bounding box supervision for improving segmentation performance.

	In this paper, we present a learning-based pseudo mask generator (LPG) specified for box-supervised semantic segmentation for producing better segmentation proposals.
	We note that the parameters of several generic proposal generators, \emph{e.g.}, MCG~\cite{MCG_proposal}, actually are obtained using an independent training set, \emph{e.g.}, BSDS500~\cite{amfm_pami2011}. 
	Instead, we resort to a pixel-level annotated auxiliary dataset where the class labels are non-overlapped with those of the box-annotated dataset.
	Considering that PASCAL VOC 2012~\cite{pascal_voc} is usually adopted in weakly supervised semantic segmentation, we use the COCO benchmark~\cite{MSCOCO} to form the auxiliary dataset by selecting the 60 object classes non-intersecting with the 20 classes in VOC 2012.
	The auxiliary dataset is leveraged to learn a class-agnostic proposal generator, which can then be deployed to any box-annotated dataset for generating accurate segmentation proposals and improving segmentation performance.
	Our LPG is of great practical value due to that: (i) Public pixel-level annotated datasets, \emph{e.g.}, PASCAL VOC \cite{pascal_voc}, MS-COCO \cite{MSCOCO}, LVIS \cite{lvis}, Open Images \cite{openimages} \emph{etc}., are ready to act as the auxiliary datasets to train LPG for future box-supervised semantic segmentation tasks. 
	(ii) In practical applications, trained LPG models can be directly applied to box-annotated datasets even with new classes and complicated scenes, without requiring any pixel-level annotations.

	We further present a bi-level optimization model to make the learned proposal generator be specified for box-supervised semantic segmentation. 
	Based on the bounding box annotations of the auxiliary dataset, we define the lower subproblem as a box-supervised semantic segmentation model. 
	With the pixel-level annotation, the upper subproblem is defined as the learning of a class-agnostic pseudo mask generator for optimizing box-supervised semantic segmentation performance.
	Then an expectation-maximization (EM) algorithm is then used for joint learning of box-supervised semantic segmentator and pseudo mask generators with multiple stages, shown in Fig. \ref{fig:figure2}.
	Moreover, stage-wise generator is adopted to cope with that the learned semantic segmentator gradually becomes more precise during training.
	By solving the bi-level optimization model, we can make the learned pseudo mask generator to be specified for optimizing box-supervised semantic segmentation performance.
	The class-agnostic setting is also beneficial to make the learned pseudo mask generator generalize well to new classes and new datasets. 
	Subsequently, the trained LPG can be deployed to boost the performance of any segmentation backbone network. 
	Finally, experiments are conducted on PASCAL VOC 2012 dataset~\cite{pascal_voc} to evaluate our method. 
	The results show that the learned pseudo mask generator is effective in boosting segmentation performance. 
	For the segmentation backbones DeepLab-LargeFOV~\cite{chen14semantic} and DeepLab-ResNet101~\cite{chen_deeplabv2}, our method outperforms the state-of-the-art approaches~\cite{BoxSup_Dai_2015,Papandreou_2015_ICCV,Khoreva_2017_SDI,BCM_Song_2019_CVPR}, and further closes the performance gap between box-supervised and fully-supervised models.

	Generally, the main contribution of this work is summarized as follows:
	\begin{itemize}
		\item A learning-based class-agnostic pseudo mask generator (LPG) is presented to produce more accurate segmentation mask specified for box-supervised semantic segmentation.
		\item A bi-level optimization model and an EM-based algorithm are proposed to learn stage-wise, class-agnostic pseudo mask generator from the auxiliary dataset. 
		\item Comprehensive experiments show that our proposed method performs favorably against state-of-the-arts, and further closes the performance gap between box-supervised and fully-supervised models.
	\end{itemize}
	
	The remainder of this paper is organized as follows: we briefly review relevant works of fully supervised and weakly supervised semantic segmentation methods in Section \ref{sec:related}. 
	In Section \ref{sec:method}, the proposed LPG is presented in details along with bi-level optimization algorithm.  
	%
	In Section \ref{sec:experiment}, experiments are conducted to verify the effectiveness of our LPG in comparison with state-of-the-art methods.
	%
	Finally, Section \ref{sec:conclusion} ends this paper with concluding remarks.

	\section{Related Work}\label{sec:related}
	In this section, we briefly survey fully supervised learning, weakly supervised learning and transfer learning for semantic segmentation.
	
	\subsection{Fully Supervised Semantic Segmentation}
	Recent success in deep learning has brought unprecedented progress in fully supervised semantic segmentation~\cite{Ross_RCNN,CRF_as_RNN,chen14semantic,Long_FCN_2015,Lin_2016_CVPR,FPN_Lin_2017,zhao2017pspnet,xiao2018unified}.
	As two representative architectures, FCN~\cite{Long_FCN_2015} and U-Net~\cite{UNetRFB15a} are very common in semantic segmentation. 
	Then advanced network modules, \emph{e.g.}, dilated convolution \cite{chen_deeplabv2}, have also been introduced to enhance the representation ability of deep networks. 
	In addition, hierarchical convolutional network~\cite{Ross_RCNN, HariharanAGM15} and pyramid architecture  ~\cite{Chen_Rethink,FPN_Lin_2017,zhao2017pspnet,xiao2018unified} have drawn much attention for enriching feature representations in semantic segmentation. 
	In PSPNet~\cite{zhao2017pspnet}, a pyramid-pooling module (PPM) is adopted with different pooling sizes.
	While in UperNet~\cite{xiao2018unified}, Xiao \emph{et al}.~introduced multi-PPM into feature pyramid networks.
	In DeepLabV3+~\cite{Chen_2018_ECCV}, depth-wise atrous convolution is adopted in encoder-decoder to effectively utilize multi-scale contextual feature information.
	HRNet~\cite{hrnet} forms parallel convolution streams to combine and exchange feature information across different resolutions.
	To better aggregate global and local features, DANet~\cite{danet_fu2019dual} suggests an additional position attention module to learn spatial relationships among features, and CCNet~\cite{ccnet_attention} replaces non-local block with a more computational efficient criss-cross attention module.
	However, these state-of-the-art fully supervised 
	approaches depends heavily on pixel-level annotations of enormous training images, severely limiting their scalability. 
	In this work, we instead focus on weakly supervised semantic segmentation. 

	\subsection{Weakly Supervised Semantic Segmentation}
	One key issue in weakly supervised approaches is how to generate proper segmentation masks from weak labels, \emph{e.g.}, image-level class labels~\cite{Ahn_2018_AffinityNet,Ahn_2019_IRNet,ClassPeak_Ins,adver_erasing,hou2018selferasing,jing2019coarse}, object points~\cite{whats_the_point,mcever2020pcams,extrem_cut}, scribbles~\cite{ScribbleSup,Regularized_loss} and bounding boxes~\cite{BoxSup_Dai_2015,Papandreou_2015_ICCV,Rajchl2017DeepCutOS,Khoreva_2017_SDI,hu2018learning,BCM_Song_2019_CVPR}.
	Among them, bounding box annotation has attracted considerable recent attention, and generic proposal generators, \emph{e.g.}, CRF, GrabCut and MCG, are usually deployed. 
	In WSSL~\cite{Papandreou_2015_ICCV}, dense CRF is adopted to enhance the estimated segmentation proposals to act as latent pseudo labels in EM algorithm for optimizing deep segmentation model.
	In BoxSup~\cite{BoxSup_Dai_2015}, MCG algorithm~\cite{MCG_proposal} is applied for producing segmentation proposals.
	In SDI~\cite{Khoreva_2017_SDI}, segmentation proposals of GrabCut~\cite{Roth_GrabCut} and MCG~\cite{MCG_proposal} are intersected to generate more accurate pseudo labels.
	Most recently, BCM \cite{BCM_Song_2019_CVPR} exploited box-driven class-wise masking and filling rate guided adaptive loss to mitigate the adverse effects of wrongly labeled proposals, while it still heavily depends on the generic proposal generator \emph{i.e.}, dense CRF.
	While Box2Seg~\cite{box2seg} introduces an additional encoder-decoder architecture with a multi-branch attention module, which further improves the segmentation performance.
	In this work, we propose a learning-based pseudo mask generator specified for generating better segmentation masks in box-supervised semantic segmentation. 
	
	\subsection{Transfer Learning for Semantic Segmentation}
	In general, synthetic datasets with pixel-level annotations are easier to collect, \emph{e.g.}, computer game dataset GTAV \cite{zhang2018fully,sun2019not} and virtual city dataset SYNTHIA \cite{sun2019not}, for fully supervised training of semantic segmentation, but they have domain gap with real-world sceneries. 
	Thus, domain adaption approaches have been studied to narrow the distribution gap between synthetic and real-world sceneries, including fully convolutional adaptation networks \cite{zhang2018fully}, hierarchical weighting networks \cite{sun2019not} and class-wise maximum squares loss \cite{chen2019domain}. 
	Moreover, Shen \emph{et al}.~\cite{shen2018bootstrapping} proposed bidirectional transfer learning to tackle a more challenging task where images from both source and target domains are with image-level annotations. 
	However, these domain adaption methods require shared class labels across domains. 
	One exception is TransferNet \cite{hong2016learning}, where class-agnostic transferable knowledge is learned from source domain with pixel-level annotations to target domain with image-level labels using encoder-decoder with an attention module.  
	But weakly supervised semantic segmentation with only image-level labels is usually inferior to that with bounding box annotations, and in \cite{hong2016learning}, a general proposal generator CRF is still adopted for boosting performance. 
	In this work, we propose a learning-based class-agnostic pseudo mask generator specified for box-supervised semantic segmentation task.

	\begin{figure*}[!t]
		\setlength{\tabcolsep}{2.0pt}
		\centering
		\includegraphics[width=0.93\linewidth]{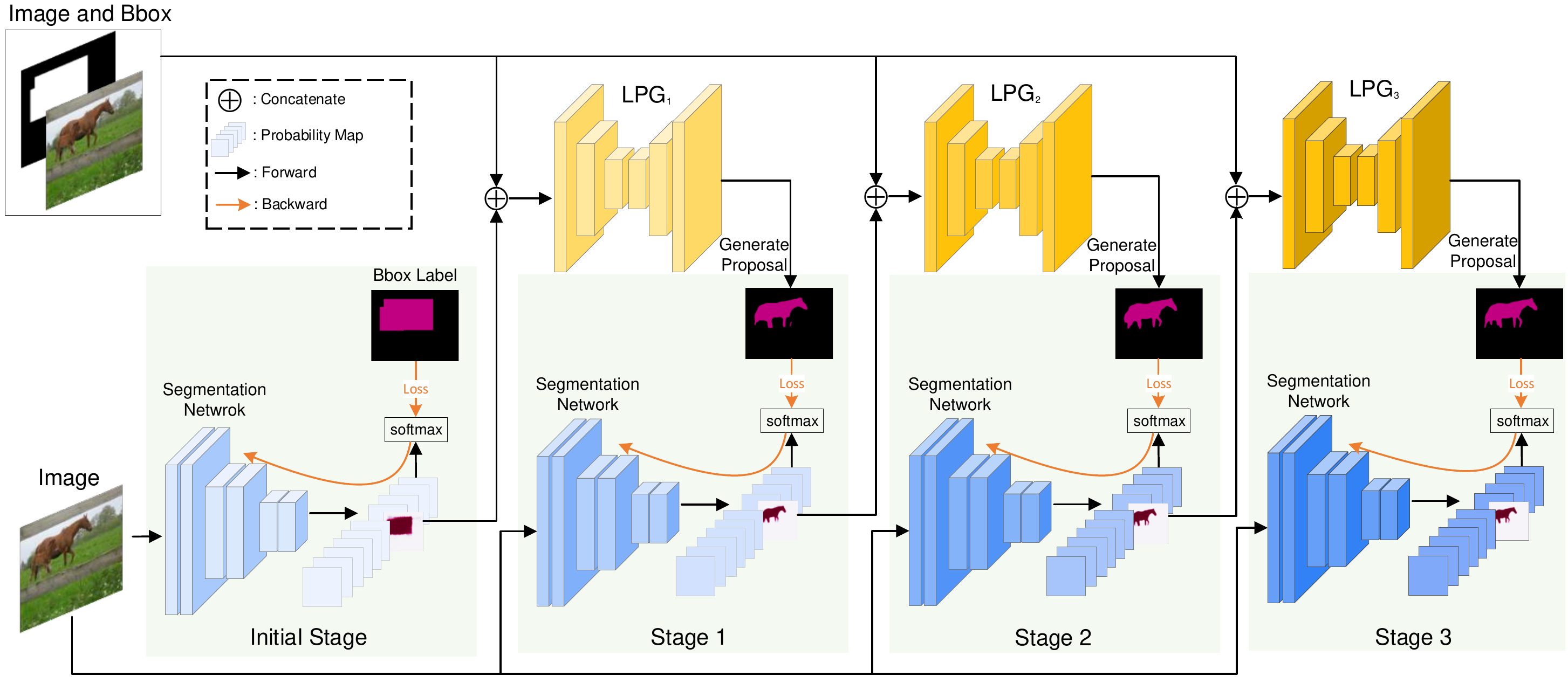}
		\caption{Training a deep segmentation network (\emph{e.g.}, DeepLab-LargeFOV~\cite{chen14semantic} or DeepLab-ResNet101~\cite{chen_deeplabv2}) on dataset $\mathcal{D}$ with only box-level annotations, where the pixel-level class-agnostic pseudo masks are generated using our LPG stage-by-stage. 
			Three LPG modules have been trained on another auxiliary dataset $\mathcal{D}_{aux}$ with pixel-level annotations, as illustrated in Fig. \ref{fig:figure2}. 
			During inference, the trained segmentation network from Stage 3 can be applied to produce satisfying segmentation results.
			In this work, we assume the semantic classes from $\mathcal{D}$ are not intersected with those from $\mathcal{D}_{aux}$ to verify the generalization ability of LPG.
		}
		\label{Overview_Fig1}
		\vspace{-1.0em}
	\end{figure*}
	
	
	\section{Proposed Method}\label{sec:method}
	In this section, we first revisit the EM-based weakly supervised semantic segmentation method, \emph{i.e.}, WSSL~\cite{Papandreou_2015_ICCV}.
	Then using the auxiliary dataset, we present our bi-level optimization formulation and the EM algorithm to learn an optimal class-agnostic pseudo mask generator.
	Finally, the learned pseudo mask generator (LPG) can be deployed to a box-annotated dataset for improving weakly-supervised semantic segmentation performance.

	In particular, we use an auxiliary dataset with both bounding box and pixel-level annotations, \emph{i.e.}, $\mathcal{D}_{aux} = \{ ({\bf x}_{aux}, {\bf y}^p_{aux}, {\bf y}^b_{aux})\}$, to train LPG.
	Here, ${\bf x}_{aux}$ denotes an image from $\mathcal{D}_{aux}$, while ${\bf y}^p_{aux}$ and ${\bf y}^b_{aux}$ respectively denote the pixel-level and bounding box annotations of ${\bf x}_{aux}$.
	Without loss of generality, we assume that the box annotation ${\bf y}^b_{aux}$ has the same form as the pixel-level annotation ${\bf y}^p_{aux}$, which can be easily attained by assigning 1 to the pixels within bounding boxes and 0 to the others for each class.
	The learned pseudo mask generator can then be used to train deep semantic segmentation model on a box-annotated dataset $\mathcal{D} = \{ ({\bf x}, {\bf y}^b)\}$.
	To verify the generalization ability of LPG, we further assume that the semantic classes from $\mathcal{D}$ are not intersected with those from $\mathcal{D}_{aux}$.
	
	\subsection{Revisit EM Algorithm in WSSL}
	In WSSL~\cite{Papandreou_2015_ICCV}, Papandreou \emph{et al}. adopted the EM algorithms for weakly supervised learning on the dataset $\mathcal{D}$.
	The pixel-level segmentation ${\bf y}^{p}$ of ${\bf x}$ is treated as latent variables.
	Denote by $\theta^{\prime}$ the previous parameters of the segmentation model, and $f({\bf x}; \theta^{\prime})$ is the predicted segmentation with the parameters $\theta^{\prime}$.
	In the EM algorithm, the E-step is used to update the latent segmentation by,
	\begin{equation}\label{E-step}
		\hat{\bf y}^{p} = \arg \max_{{\bf y}^{p}} \log P({\bf y}^{b} |{\bf y}^{p}) - \mathcal{L}({\bf y}^{p}, f({\bf x}; \theta^{\prime})),
	\end{equation}
	where $\mathcal{L}({\bf y}^{p}, f({\bf x}; \theta^{\prime}))$ usually is the cross-entropy loss.
	Then, the M-step is deployed to update the model parameter $\theta$,
	\begin{equation}\label{M-step}
		{\theta} = \arg \min_{\theta} \mathcal{L}(\hat{\bf y}^{p}, f({\bf x}; \theta)).
	\end{equation}
	For image-level weak supervision, WSSL~\cite{Papandreou_2015_ICCV} suggests both EM-Fixed and EM-Adapt methods in the E-step.
	For box supervision, WSSL~\cite{Papandreou_2015_ICCV} further considers Bbox-Seg to exploit dense CRF~\cite{DenseCRF} for generating segmentation proposal.
	In~\cite{BoxSup_Dai_2015,Khoreva_2017_SDI}, MCG and GrabCut are used to iteratively generating proposals for training segmentation models, which can also be regarded as EM-like procedures.
	
	\subsection{Learning-based Pseudo Mask Generator}
	\textbf{Problem Formulation.} We note that either EM-Fixed, EM-Adapt, or dense CRF, MCG, GrabCut are generic and not designed for directly optimizing weakly supervised semantic segmentation performance.
	Thus, instead of generic proposal generators, we aim at learning an optimized pseudo mask generator from $\mathcal{D}_{aux}$ by solving a bi-level optimization model.
	To this end, we suggest to utilize a class-agnostic pseudo mask generation network, \emph{i.e.}, $g({\bf y}^{b}_{aux}, {\bf x}_{aux}, f({\bf x}_{aux}; \theta^{\prime}); \omega)$, to update the latent segmentation on $\mathcal{D}_{aux}$,
	\begin{equation}\label{E-step1}
		\hat{\bf y}^{p}_{aux} = g({\bf y}^{b}_{aux}, {\bf x}_{aux}, f({\bf x}_{aux}; \theta^{\prime}); \omega),
	\end{equation}
	where $\omega$ is the parameters of the pseudo mask generation network.
	By substituting the above equation into the M-step, we have the lower subproblem on $\theta$,
	\begin{equation}\label{lower0}
		\hat{\theta} = \arg \min_{\theta} \mathcal{L}(\hat{\mathbf{y}}_{aux}^p, f({\bf x}_{aux}; \theta)).
	\end{equation}
	Taking the above two equations into consideration, the learned parameters $\hat{\theta}$ is a function on $\omega$, \emph{i.e.}, $\hat{\theta}(\omega)$, and we have the pixel-level annotations on $\mathcal{D}_{aux}$.
	Thus, the upper subproblem on $\omega$ can be defined for optimizing the performance of the box-supervised semantic segmentation model $f({\bf x}_{aux}; \hat{\theta})$,
	\begin{equation}\label{upper0}
		\omega = \arg \min_{\omega} \mathcal{L}({\bf y}_{aux}^p, f({\bf x}_{aux}; \hat{\theta}(\omega))).
	\end{equation}
	When the pseudo mask generator produce the ground-truth segmentation masks, the M-step will be a fully-supervised setting, and thus the gap between box and fully supervised learning can be completely eliminated.
	Thus, to ease the training difficulty, we modify the upper subproblem by requiring the pseudo mask generator accurately predicts the ground-truth segmentation, resulting the following bi-level optimization formulation,
	\begin{equation}\label{upper}
		\begin{aligned}
			\omega = \arg \min_{\omega} &\mathcal{L}({\bf y}_{aux}^p, g({\bf y}^{b}_{aux}, {\bf x}_{aux}, f({\bf x}_{aux}; \hat{\theta}); \omega)),\\
			\mbox{s.t. } \hat{\theta} =& \arg \min_{\theta} \mathcal{L}(\hat{\mathbf{y}}_{aux}^p, f({\bf x}_{aux}; \theta)),
		\end{aligned}
	\end{equation}
	where the cross-entropy loss is adopted as $\mathcal{L}$ in both upper and lower optimization problems.

	\textbf{Network Architectures.} The training on $\mathcal{D}_{aux}$ involves both the pseudo mask generation network and the deep segmentation network.
	For the deep segmentation network, we keep consistent with most existing weakly supervised semantic segmentation methods, and adopt DeepLab-LargeFOV~\cite{chen14semantic} or DeepLab-ResNet101~\cite{chen_deeplabv2} as the backbone. 
	As for the pseudo mask generation network, we adopt the Hourglass~\cite{hourglassNet} structure. 
	It takes the input image, bounding box map ${\bf y}^{b}_{aux}$ for a specific class and the predicted segmentation for a specific class as the input to generate the segmentation mask of the corresponding class.
	Concretely, the output of the proposal generation network consists of two channels, with one for estimating the object mask inside the bounding box and the other for the remaining unrelated background.

	\textbf{Learning Pseudo Mask Generation Network.} We further present an EM-like algorithm, as shown in Fig. \ref{fig:figure2}, to alternate between updating the pseudo mask generation network and the deep segmentation network. 
	To begin with, we first adopt ${\bf y}^{b}_{aux}$ for the initialization $\hat{\bf y}^{p}_{aux}$, which is then used to train the deep segmentation network $f({\bf x}_{aux}; \theta)$. 
	In the E-step, the pseudo mask generation network takes the input image ${\bf x}_{aux}$, bounding box annotation $\hat{\bf y}^{b}_{aux}$ and the updated $f({\bf x}_{aux}; \theta)$ as the input.
	And the pixel level annotation is adopted as the ground-truth to train the network.   
	In the M-step, the deep segmentation network takes an image ${\bf x}_{aux}$ as the input.
	And the segmentation proposals generated by $g({\bf y}^{b}_{aux}, {\bf x}, f({\bf x}_{aux}; \theta^{\prime}); \omega)$ is used as the ground-truth for supervising network training.
	We also note that the input of the pseudo mask generation network changes during training. 
	Along with the updating of deep segmentation network, the predicted segmentation $f({\bf x}_{aux}; \theta^{\prime})$ becomes more and more accurate. 
	Consequently, the final pseudo mask generation network may only be applicable to the last stage of the deep segmentation network training, and is inappropriate for generating precise pseudo masks for the early training of deep segmentation network.
	As a remedy, we treat each round of the E-step and M-step as a stage and suggest to learn stage-wise pseudo mask generator. 
	We empirically find that three or four stages are sufficient in training.

	\textbf{Discussion.} Benefited from the bi-level optimization formulation and class-agnostic setting, our learning-based pseudo mask generator can be optimized for box-supervised semantic segmentation and is able to generalize well to new classes and new datasets.
	Generally, it is believed that more accurate pseudo mask generator is beneficial to the segmentation performance.
	In Eqn.~(\ref{upper}), the pseudo mask generation network takes the box annotation as the input, and is trained to generate pseudo masks approaching the ground-truth.
	Thus, it is specified for box supervision and can be used to boost semantic segmentation performance.
	Moreover, the pseudo mask generation network is class-agnostic.  
	For any given semantic class, we use the same pseudo mask generation network to generate a segmentation mask for this class.
	Thus, it exhibits good generalization ability and can generalize well to new classes. 
	
	One may notice that TransferNet \cite{hong2016learning} adopts a similar setting with ours, where class-agnostic transferable knowledge is learned from auxiliary dataset with pixel-level annotations to another dataset with image-level annotations.   
	Our LPG differs from \cite{hong2016learning}: 
	(i) LPG aims to generate pixel-level segmentation masks from bounding boxes, which provide certain object localization information to guarantee the superior performance of weakly semantic segmentation. 
	(ii) LPG is completely learned for better generating accurate pixel-level pseudo masks, while in \cite{hong2016learning} the proposals are obtained using an attention model that also requires to adopt generic proposal generator dense CRF for boosting segmentation performance.

	\subsection{Box-Supervised Semantic Segmentation}
	Owing to its optimized performance and generalization ability, the learned pseudo mask generator can be readily deployed to any box-annotated dataset $\mathcal{D}$ for weakly supervised semantic segmentation, as shown in Fig. \ref{Overview_Fig1}.
	Analogous to the EM-like algorithm, we first adopt ${\bf y}^{b}$ for training the deep semantic segmentation network $f({\bf x}; \theta)$. 
	Then, the learned pseudo mask generator in the first stage is used to generate more accurate segmentation mask, \emph{i.e.}, $g({\bf y}^{b}, {\bf x}, f({\bf x}; \theta); \omega)$, which is then utilized for further training the deep semantic segmentation network.
	Subsequently, the learned pseudo mask generator in the second and later stages can be deployed.  
	Finally, the trained semantic segmentation network from the last stage can be applied to produce satisfying segmentation results. 
	For verifying the generalization ability of the learned pseudo mask generator, we assume that
	the semantic classes from $\mathcal{D}$ are not intersected with those from $\mathcal{D}_{aux}$.
	Undoubtedly, the learned pseudo mask generator should work better when the semantic classes from $\mathcal{D}$ and $\mathcal{D}_{aux}$ are overlapped.
	
	\subsection{Generating Pseudo Masks with Multi-Classes}
	%
	Finally, we discuss how to generate segmentation masks using LPG model for an image that contains multi-classes of objects.
	Assuming that there are total $N$ categories of objects in the target dataset $\mathcal{D}$ (\emph{i.e.}, $N=20$ for PASCAL VOC 2012~\cite{pascal_voc}). 
	Given a stage, the LPG model $g$ with parameters $\theta$ takes box annotations $\mathbf{y}^b$, input image $\mathbf{x}$ and current segmentation results $f(\mathbf{x};\theta^{\prime})$ as input, and its output is normalized by \emph{softmax} function. 
	The multi-classes masks $\mathbf{y}^p$ are generated by 
	\begin{equation}\label{multi-cat-obj}
		\begin{aligned}
			\mathbf{p} = & \textit{softmax}(g({\bf y}^{b}, {\bf x}, f({\bf x}; \theta^{\prime}); \omega)),\\
			{\bf y}^{p} = & \sum_{C_{i=1}}^{N} C_{i} \times {\bf y}^{b_{i}} \bullet \psi(\mathbf{p}) ,
		\end{aligned}
	\end{equation}
	where ${\bf y}^{b_{i}}$ is the bounding box annotation of objects from $i$-th class, $C_{i}$ is the $i$-th class label and $\bullet$ is the entry-wise product. 
	Since the output of LPG is a two-channel probability map, we adopt a $\psi(\mathbf{p})$ function to extract the binary mask from $\mathbf{p}$, where 1 indicates the foreground and 0 denotes the background. 
	We simply implement $\psi$ as the \emph{numpy.argmax}~\cite{numpy_harris2020array} function in Python.
	For overlapping objects from two classes, we calculate $\text{overlapping rate} = \frac{\text{overlapping area}}{\text{predicted semantic area}}$, and assign this region to the class with higher overlapping rate.

	\section{Experiments}\label{sec:experiment}
	In this section, we first discuss the effectiveness of stage-wise LPG for generating pseudo masks and the architecture of LPG network. 
	Then, our method is compared with state-of-the-art box-supervised semantic segmentation methods in both weakly supervised and semi-supervised manners. 
	For evaluating the segmentation performance, mean pixel Intersection-over-Union (mIoU) is adopted as the quantitative metric. 
	Our source code and all the pre-trained models have been made publicly available at \url{https://github.com/Vious/LPG_BBox_Segmentation}.

	\subsection{Experimental Setting}
	\textbf{Dataset.} 
	\quad 
	In weakly supervised manner, we follow \cite{BCM_Song_2019_CVPR,BoxSup_Dai_2015,Papandreou_2015_ICCV,Khoreva_2017_SDI} to deploy PASCAL VOC 2012 \cite{pascal_voc} dataset as $\mathcal{D}$, which contains 1,464 images in training set and 1,449 images in validation set. 
	Following \cite{Semantic_TrainAug,Khoreva_2017_SDI,Papandreou_2015_ICCV}, the training set is augmented to have 10,582 images for training segmentation networks, while 1,449 validation images are adopted for evaluating their performance.
	To train our LPG, we choose MS-COCO dataset~\cite{MSCOCO} as the auxiliary dataset $\mathcal{D}_{aux}$.  
	MS-COCO contains 80 object classes, among which 20 classes appear in PASCAL VOC 2012. 
	To verify the generalization ability of class-agnostic LPG, we exclude these 20 classes from $\mathcal{D}_{aux}$.   
	In semi-supervised manner, the only difference is in training segmentation networks on PASCAL VOC 2012, where the original 1,464 training images are provided with pixel-level annotations, while the other augmented 9,118 training images are provided with box-level annotations.  
	
	\begin{figure*}[hbt]
		\small
		\setlength{\tabcolsep}{2.0pt}
		\centering
		\begin{tabular}{ccccccc}
			\includegraphics[width=23mm,height=17mm]{Proposals/Image/2007_001825} &
			\includegraphics[width=23mm,height=17mm]{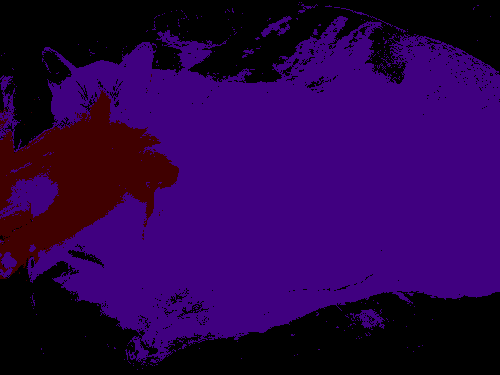} &
			\includegraphics[width=23mm,height=17mm]{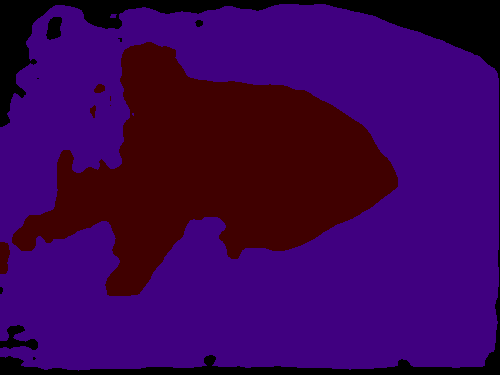} &
			\includegraphics[width=23mm,height=17mm]{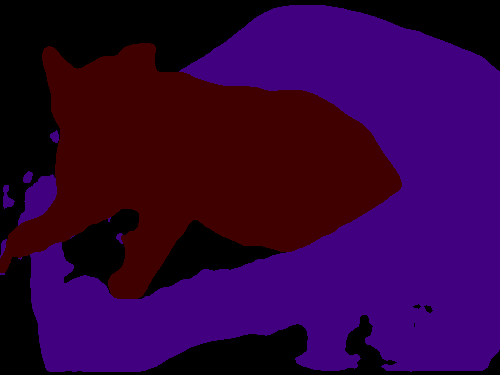} &
			\includegraphics[width=23mm,height=17mm]{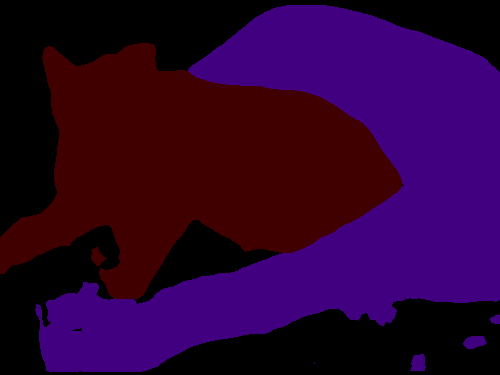} &
			\includegraphics[width=23mm,height=17mm]{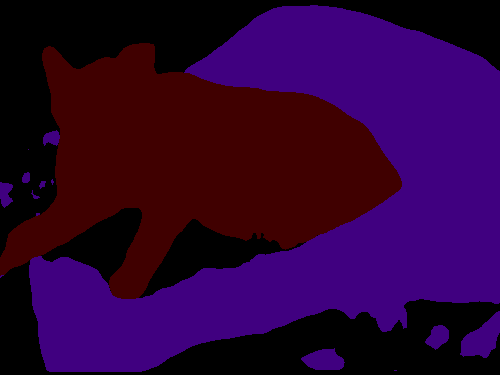} &
			\includegraphics[width=23mm,height=17mm]{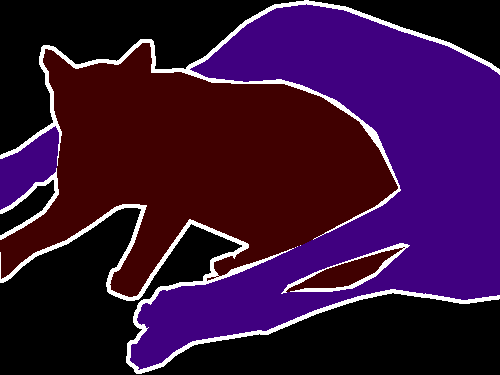}
			\\
			
			\includegraphics[width=23mm,height=17mm]{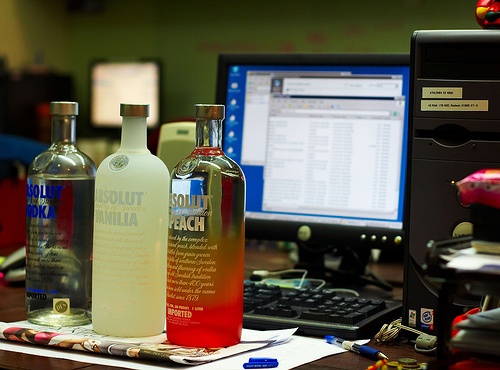} &
			\includegraphics[width=23mm,height=17mm]{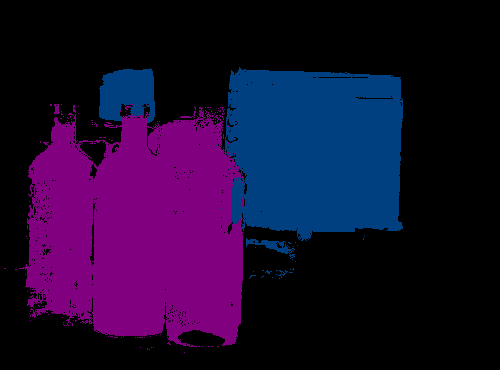} &
			\includegraphics[width=23mm,height=17mm]{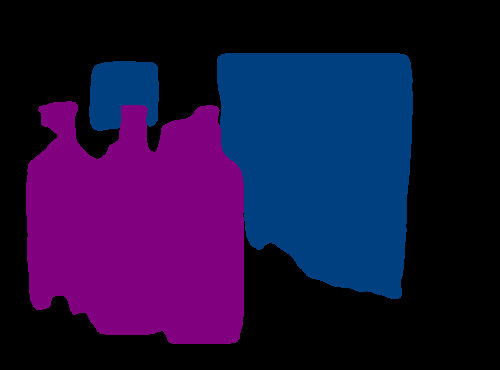} &
			\includegraphics[width=23mm,height=17mm]{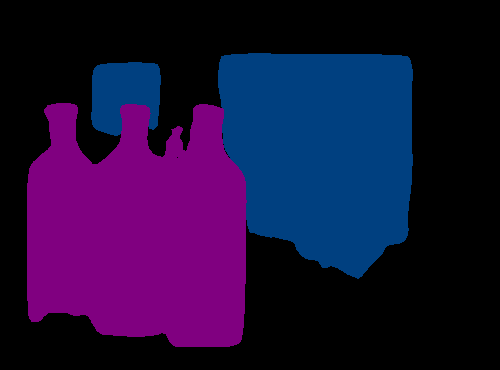} &
			\includegraphics[width=23mm,height=17mm]{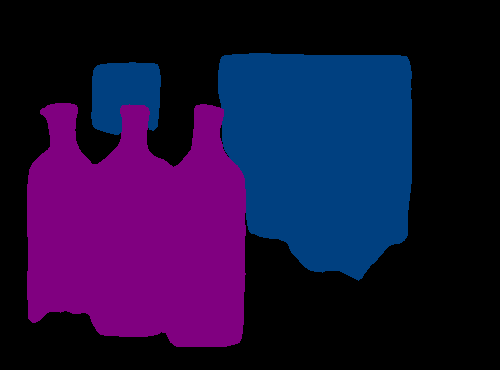} &
			\includegraphics[width=23mm,height=17mm]{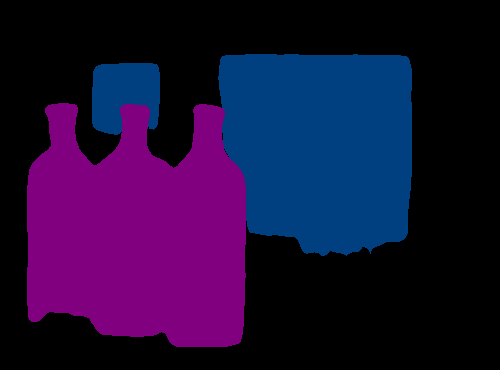} &
			\includegraphics[width=23mm,height=17mm]{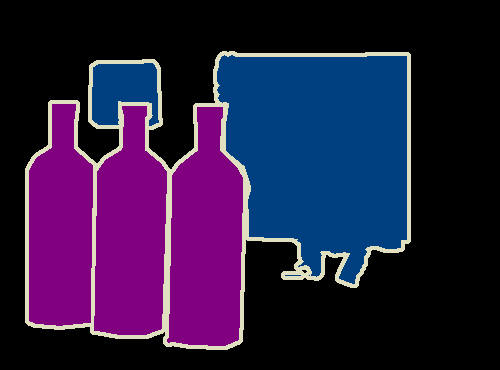}
			\\
			
			\includegraphics[width=23mm,height=17mm]{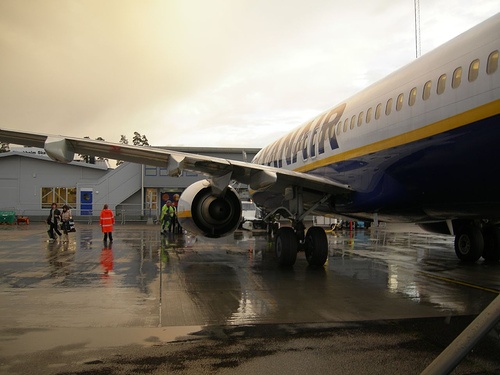} &
			\includegraphics[width=23mm,height=17mm]{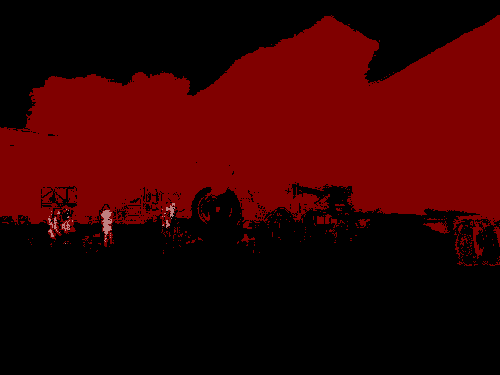} &
			\includegraphics[width=23mm,height=17mm]{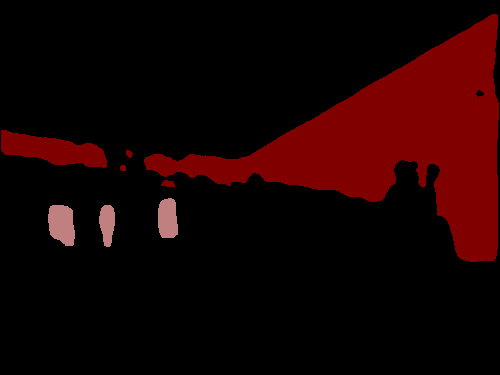} &
			\includegraphics[width=23mm,height=17mm]{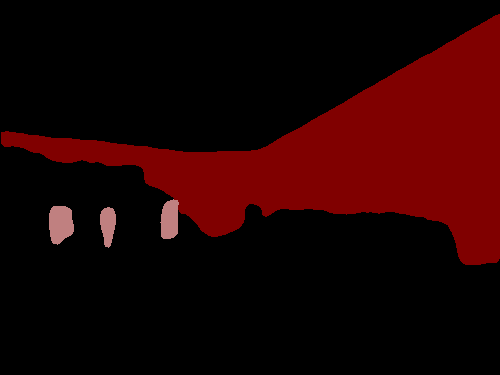} &
			\includegraphics[width=23mm,height=17mm]{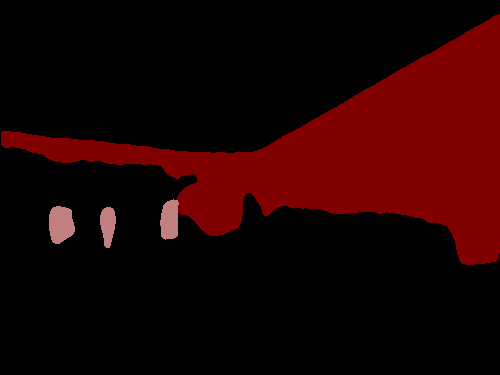} &
			\includegraphics[width=23mm,height=17mm]{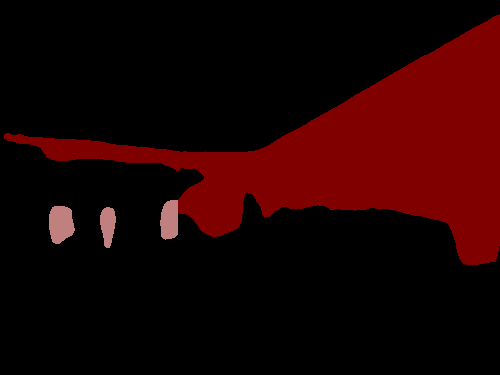} &
			\includegraphics[width=23mm,height=17mm]{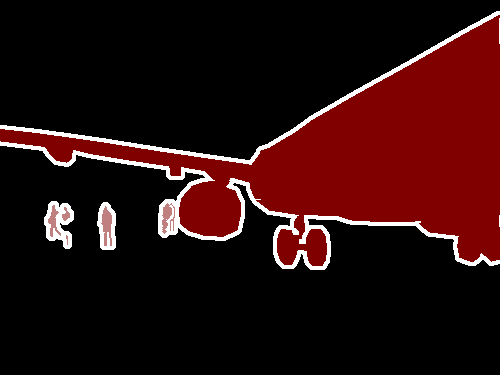}
			\\
			
			\includegraphics[width=23mm,height=17mm]{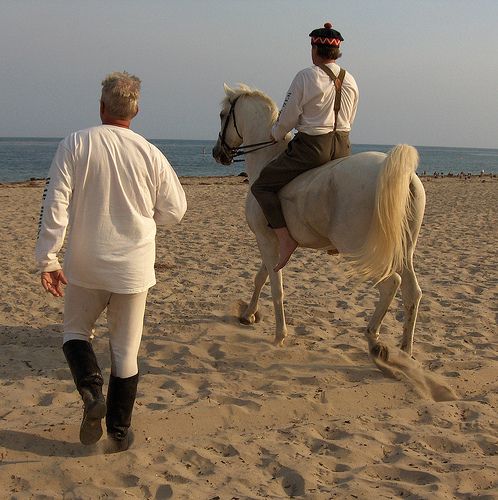} &
			\includegraphics[width=23mm,height=17mm]{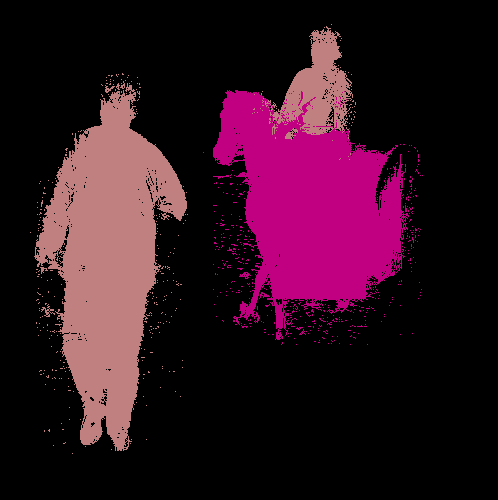} &
			\includegraphics[width=23mm,height=17mm]{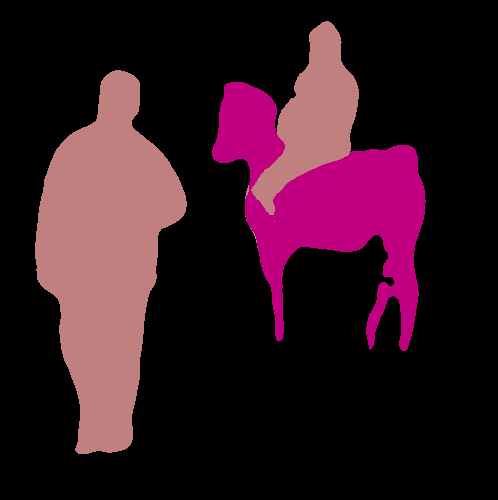} &
			\includegraphics[width=23mm,height=17mm]{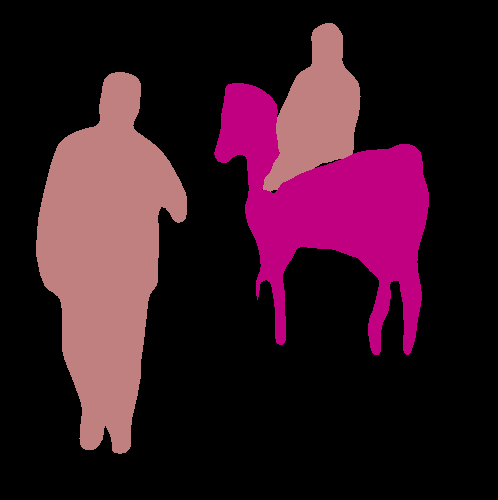} &
			\includegraphics[width=23mm,height=17mm]{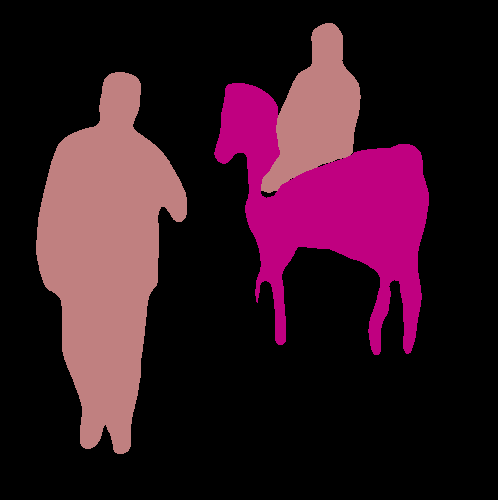} &
			\includegraphics[width=23mm,height=17mm]{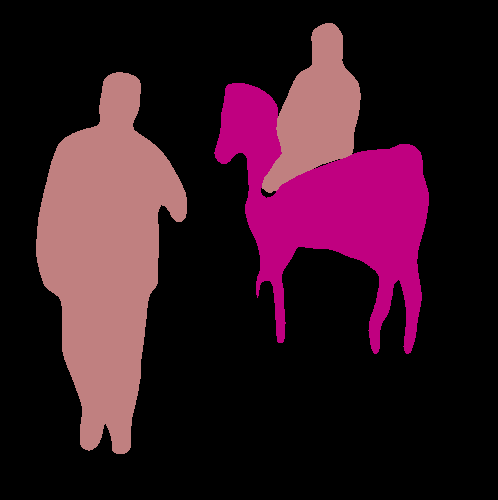} &
			\includegraphics[width=23mm,height=17mm]{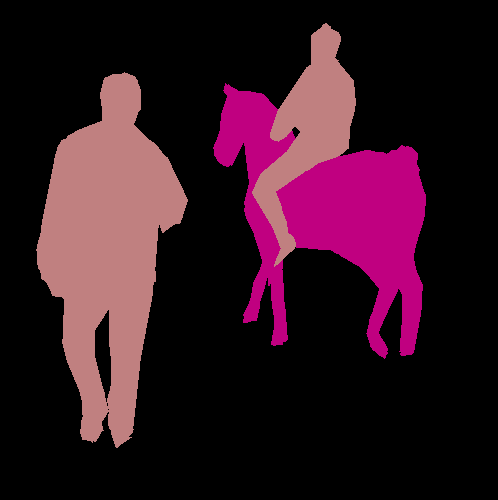}
			\\

			Images & CRF & LPG (Stage 1) & LPG (Stage 2) & LPG (Stage 3) & LPG (Stage 4) & Ground-truth \\

		\end{tabular}
		\caption{Comparison of segmentation masks generated by LPG from different stages, where DeepLab-ResNet101~\cite{chen_deeplabv2} is adopted for segmentation backbone and HourglassNet \cite{hourglassNet} is adopted for LPG. }
		\label{fig:proposals}
		\vspace{-1.5em}
	\end{figure*}

	\begin{table}[hbt]
		\centering
		\caption{Settings of hyper-parameters for training LPG and segmentation backbone.}
		\label{tabel-details}
		\resizebox{0.48\textwidth}{!}
		{
			\begin{tabular}{c|c|c|c|c|ccc}
				\hline
				
				\hline
				Model & Part & Initial $lr$ & $Max_{iter}$ & Batch size & Optimizer \\
				\hline
				
				\hline
				LPG & Full & $1.0\times10^{-4}$ &  $1\times10^5$ & 32 & Adam \\
				\hline
				
				\hline
				\multirow{4}*{LargeFOV} & $head_{\text{params}}$ & $1.0\times 10^{-3}$ & \multirow{4}*{$2\times10^4$} & \multirow{4}*{30} & \multirow{4}*{SGD} \\
				\cline{2-3}
				& $head_{\text{bias}}$ & $2.0\times10^{-3}$ &  & & \\
				\cline{2-3}
				& $last_{\text{params}}$ & $1.0\times10^{-2}$  & & & \\
				\cline{2-3}
				& $last_{\text{bias}}$ & $2.0\times10^{-2}$  & & & \\
				\hline
				
				\hline
				\multirow{4}*{ResNet101} & $head_{\text{params}}$ & $2.5\times10^{-4}$ & \multirow{4}*{$4\times10^4$} & \multirow{4}*{10} & \multirow{4}*{SGD} \\
				\cline{2-3}
				& $head_{\text{bias}}$ & $5.0\times10^{-4}$  & & & \\
				\cline{2-3}
				& $last_{\text{params}}$ & $2.5\times10^{-3}$  & & & \\
				\cline{2-3}
				& $last_{\text{bias}}$ & $5.0\times10^{-3}$ & & & \\
				\hline
				
				\hline
				
		\end{tabular}}
		\vspace{-1.0em}
	\end{table}
	\textbf{Training Details.}
	\quad 
	In our experiments, DeepLab-LargeFOV~\cite{chen14semantic} and DeepLab-ResNet101~\cite{chen_deeplabv2} are deployed as segmentation backbone to verify the effectiveness of LPG, keeping consistent with most existing weakly supervised semantic segmentation methods \cite{Khoreva_2017_SDI,Papandreou_2015_ICCV,BCM_Song_2019_CVPR}. 
	The parameters of DeepLab-LargeFOV~\cite{chen14semantic} and DeepLab-ResNet101~\cite{chen_deeplabv2} are initialized from VGG16~\cite{SimonyanZ14a} and ResNet101~\cite{resnet_he} networks trained on ImageNet~\cite{ILSVRC15} for classification. 
	The training procedures of LPG and segmentation networks are carried out using Pytorch~\cite{pytorch} on two NVIDIA GTX $2080$Ti GPUs. 
	
	For training LPG on the auxiliary dataset MS-COCO, the initial learning rate is set at $1.0\times 10^{-4}$ for each training stage, and Adam~\cite{AdamOPtim} algorithm with $\beta=(0.9, 0.999)$ is adopted for training LPG. 
	As for training segmentation backbone on both MS-COCO and PASCAL VOC 2012, we adopt SGD as optimizer and follow the parameter settings suggested by DeepLab~\cite{chen_deeplabv2}. 
	In particular, different learning rates are set for the four parts of a segmentation backbone, \emph{i.e.}, (\romannumeral1) $last_\text{bias}$: bias of the last layer, (\romannumeral2) $last_\text{params}$: parameters except bias of the last layer, (\romannumeral3) $head_\text{bias}$: bias of non-last layers, (\romannumeral4) $head_\text{params}$: parameters except bias of non-last layers.  
	The hyper-parameters for training these four parts are listed in Table~\ref{tabel-details}, where $Max_{iter}$ is the maximum training iterations.
	Moreover, we adopt the Poly~\cite{chen_deeplabv2} strategy to adjust learning rates during training all the models in each stage, which decreases an initial learning rate $lr$ by $lr = lr \times (1 - \frac{iter}{Max_{iter}} )^{0.9}$.

	\begin{table}[hbt]
		\centering
		\caption{Effect of the number of stages and network architecture of LPG. 
			Semantic segmentation backbone is adopted as DeepLab-LargeFOV~\cite{chen14semantic}, and mIoU is reported on validation set of PASCAL VOC 2012 \cite{pascal_voc}.}
		\label{ablation}
		\small
		\resizebox{0.48\textwidth}{!}{
			\begin{tabular}{c|cccc|c}
				\hline
				
				\hline  
				LPG Model  & Stage 1 & Stage 2 & Stage 3 & Stage 4 &\#Params \\
				\hline  
				ResNet18 & 64.7 & 65.9 & 66.5 & 66.6 &11.1M \\
				\hline
				ResNet101 & 65.6 & 66.6 & 67.1 & 67.2 &42.5M \\
				\hline
				HgNet (Fixed) & 64.6 & 66.4 & 67.5 & 67.6 & 12.1M \\
				\hline
				HgNet (EM)& 66.8 & 68.3 & 68.4 & 68.3 & 12.1M \\
				\hline
				
				\hline
		\end{tabular}}
		\vspace{-1.0em}
	\end{table}

	\begin{table}[!t]\small
		\vspace{-0.6em}
		\centering
		\caption{Mean IoU comparison of pseudo masks generated by stage-wise LPG and dense-CRF on PASCAL VOC 2012. }
		\label{LPG-performance}
		\begin{tabular}{c|c|c|c|c}
			\hline
			
			\hline
			
			& Stage-1 & Stage-2 & Stage-3 & CRF[23] \\
			\hline
			LargeFOV &78.2 & 81.7 & 82.2 & \multirow{2}*{69.7} \\
			\cline{1-4}
			ResNet101 &78.9 & 82.3 & 82.6 & \\
			
			\hline
			
			\hline
		\end{tabular}
		\vspace{-1.0em}
	\end{table}
	
	\subsection{Ablation Study}
	By adopting DeepLab-LargeFOV \cite{chen14semantic} as semantic segmentation backbone, we discuss how to choose the number of stages and the architecture of LPG.
	
	\textbf{Stage-wise Pseudo Mask Generators.}\quad
	By adopting three networks, \emph{i.e.}, ResNet18~\cite{resnet_he}, ResNet101~\cite{resnet_he} and HourglassNet (HgNet)~\cite{hourglassNet}, as the architecture of LPG, we discuss the number of stages. 
	As reported in Table \ref{ablation}, the segmentation performance of all the three models is improved when increasing the number of stages. 
	The performance gains from Stage 1 to Stage 3 are very significant, while the LPG by adding Stage 4 is only marginally better or even performs slightly inferior to that with three stages.
	From Fig. \ref{fig:proposals}, the generated masks become more accurate along with increasing LPG stages, and there is little difference between Stage 3 and Stage 4. 
	Thus, we suggest to set the number of stages as 3 to balance the segmentation performance and training time. 
	
	\textbf{Fixed LPG.}\quad
	To further illustrate the importance of stage-wise training, we have compared stage-wise LPG with the fixed final LPG for training Deeplab-LargeFOV~\cite{chen14semantic} on VOC 2012. 
	From Table \ref{ablation}, one can see that the final LPG (HgNet-Fixed) is not the optimal choice for early stages, and segmentation results by stage-wise LPG are consistently better than fixed LPG for any stage. 
	
	\textbf{LPG Architecture.}\quad
	As for the architecture of LPG, it is reasonable to see ResNet101~\cite{resnet_he} with more parameters is a better choice than ResNet18~\cite{resnet_he}.
	Interestingly, HourglassNet~\cite{hourglassNet} with less parameters is superior to ResNet101.
	The reason may be attributed to that the multi-scale architecture in HourglassNet can better extract the pixel-level segmentation masks. 
	Therefore, we choose HourglassNet~\cite{hourglassNet} as the default architecture of LPG.

	\begin{figure*}[!ht]
		\small
		\setlength{\tabcolsep}{2.0pt}
		\centering
		\begin{tabular}{cccccc}
			(a) & 
			\includegraphics[width=32mm,height=21mm]{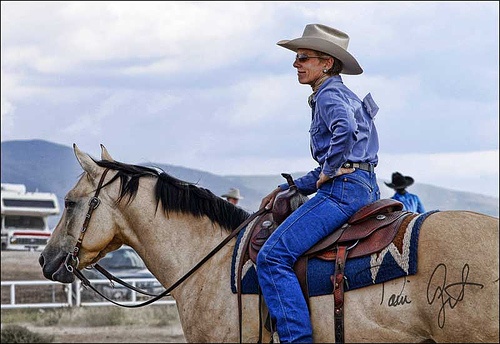} &
			\includegraphics[width=32mm,height=21mm]{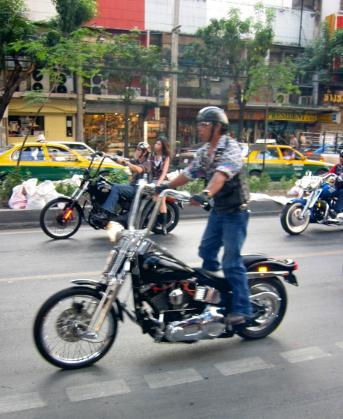} &
			\includegraphics[width=32mm,height=21mm]{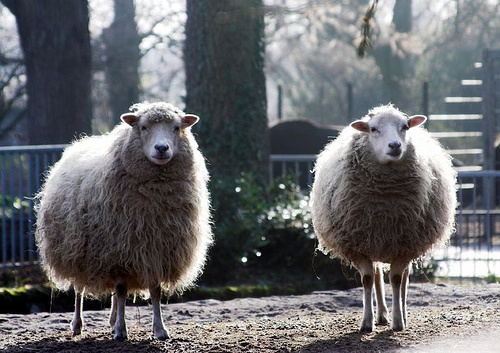} &
			\includegraphics[width=32mm,height=21mm]{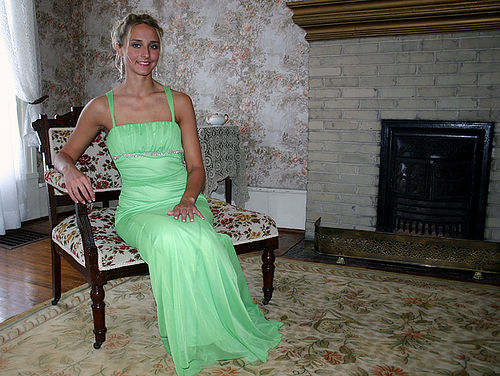} &
			\includegraphics[width=32mm,height=21mm]{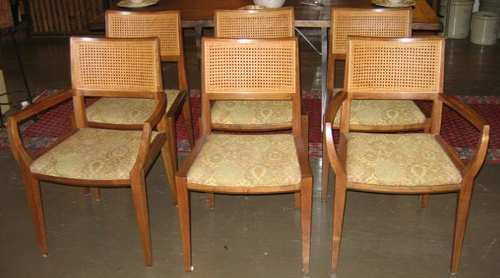} 
			\\
			
			(b)& 
			\includegraphics[width=32mm,height=21mm]{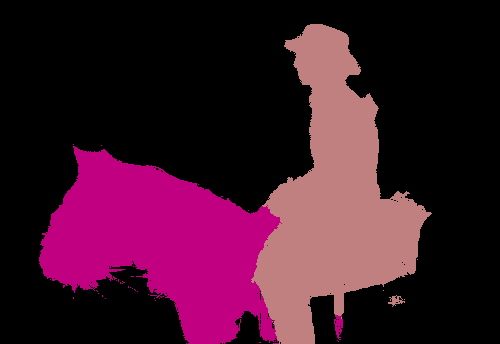} &
			\includegraphics[width=32mm,height=21mm]{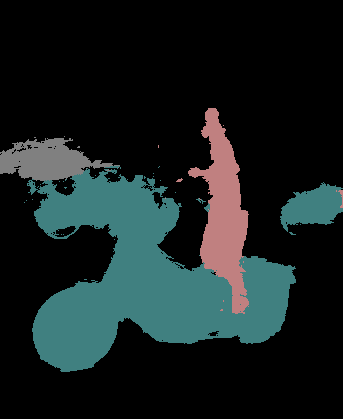} &
			\includegraphics[width=32mm,height=21mm]{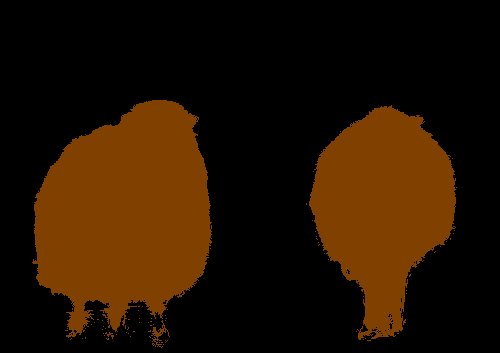} &
			\includegraphics[width=32mm,height=21mm]{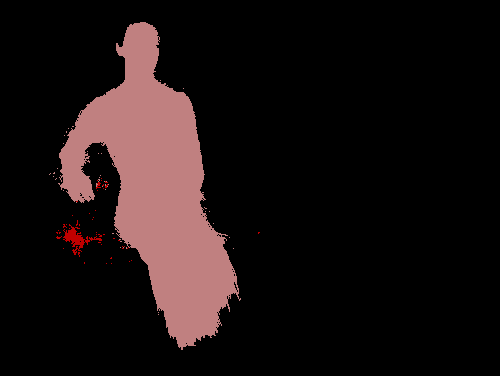} &
			\includegraphics[width=32mm,height=21mm]{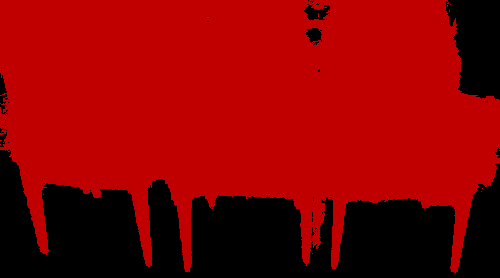} \\

			(c) &
			\includegraphics[width=32mm,height=21mm]{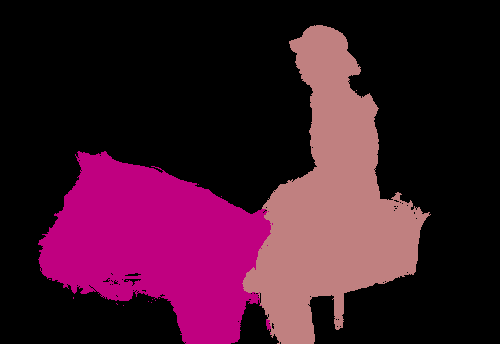} &
			\includegraphics[width=32mm,height=21mm]{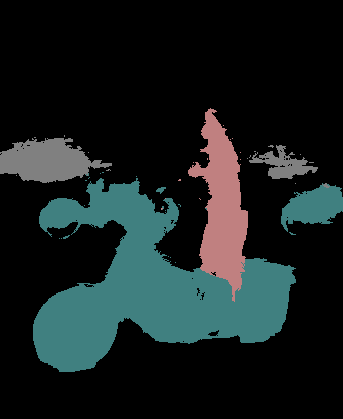} &
			\includegraphics[width=32mm,height=21mm]{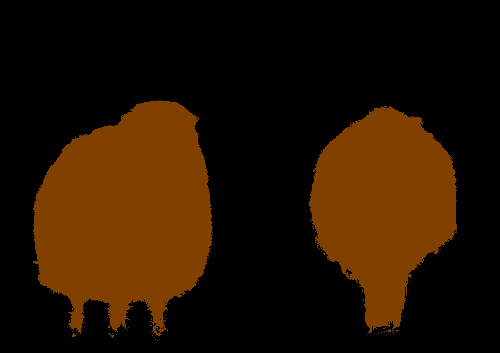} &
			\includegraphics[width=32mm,height=21mm]{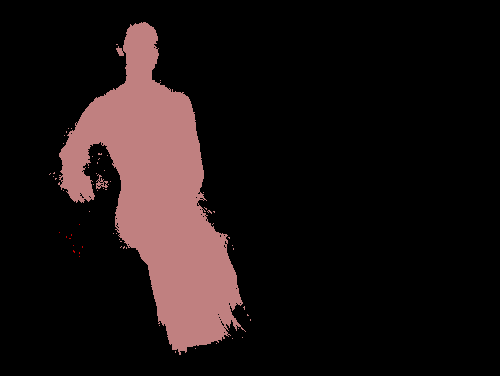} &
			\includegraphics[width=32mm,height=21mm]{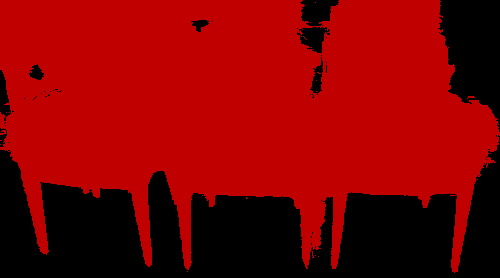} \\
			
			(d) &
			\includegraphics[width=32mm,height=21mm]{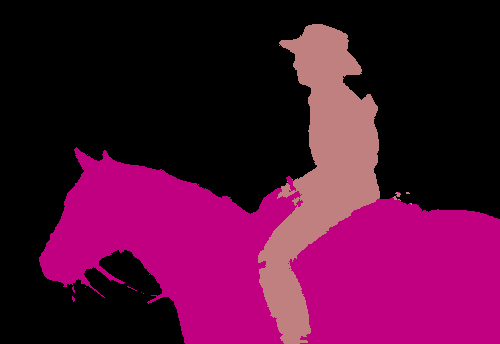} &
			\includegraphics[width=32mm,height=21mm]{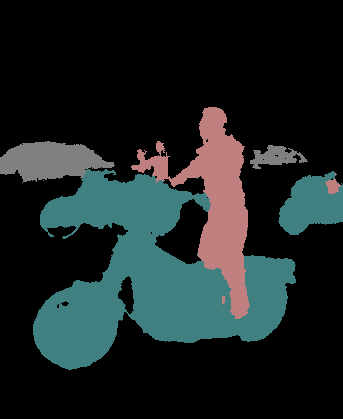} &
			\includegraphics[width=32mm,height=21mm]{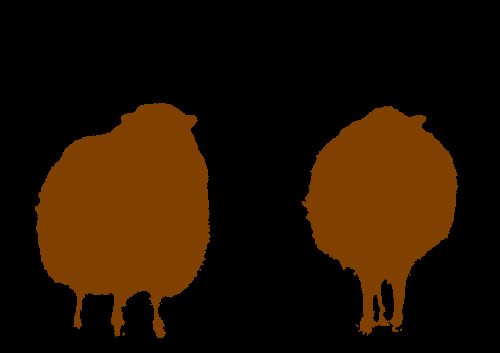} &
			\includegraphics[width=32mm,height=21mm]{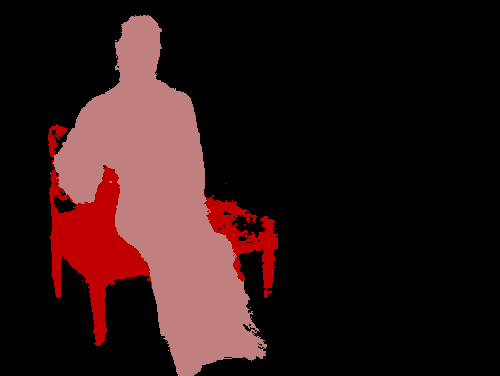} &
			\includegraphics[width=32mm,height=21mm]{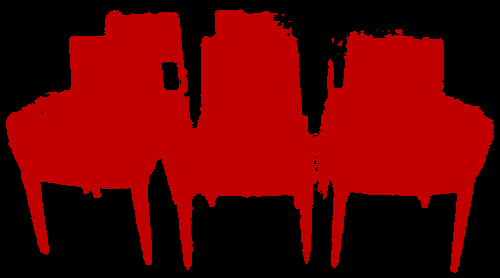} \\
			
			(e) &
			\includegraphics[width=32mm,height=21mm]{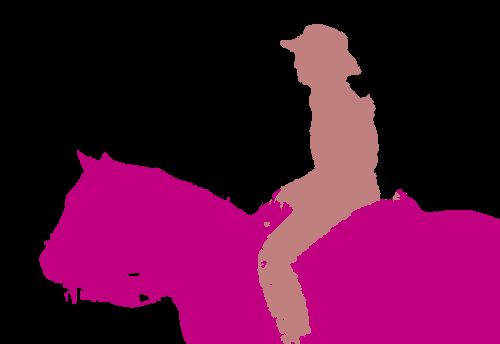} &
			\includegraphics[width=32mm,height=21mm]{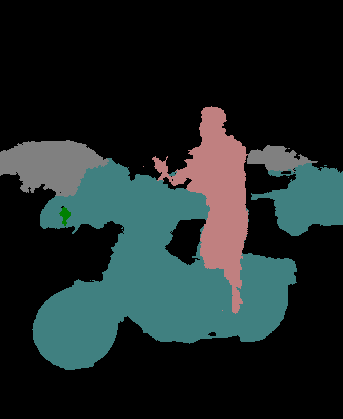} &
			\includegraphics[width=32mm,height=21mm]{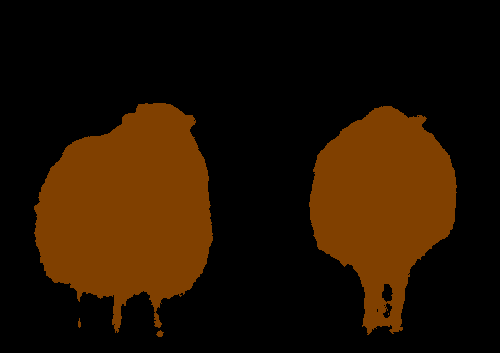} &
			\includegraphics[width=32mm,height=21mm]{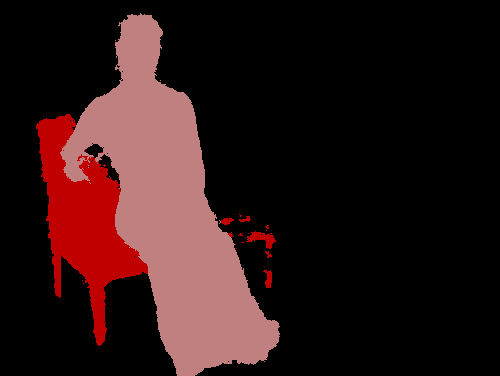} &
			\includegraphics[width=32mm,height=21mm]{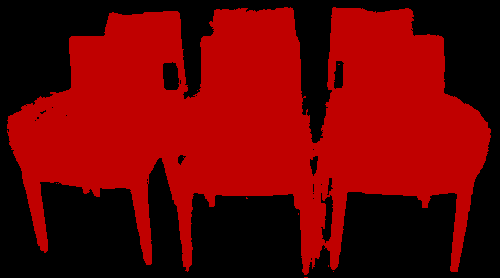} \\
			
			(f) & 
			\includegraphics[width=32mm,height=21mm]{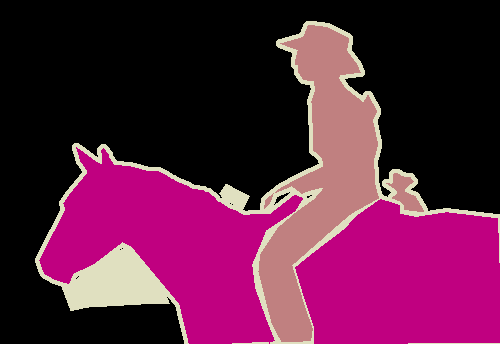} &
			\includegraphics[width=32mm,height=21mm]{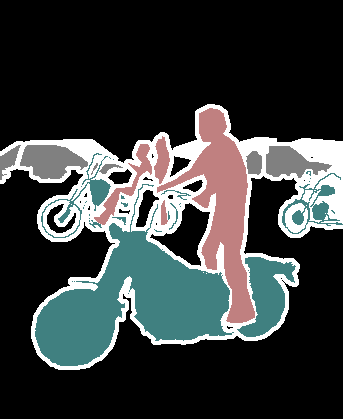} &
			\includegraphics[width=32mm,height=21mm]{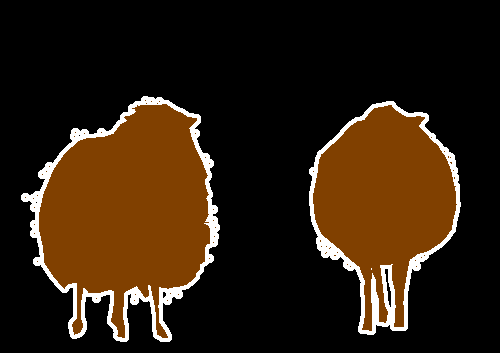} &
			\includegraphics[width=32mm,height=21mm]{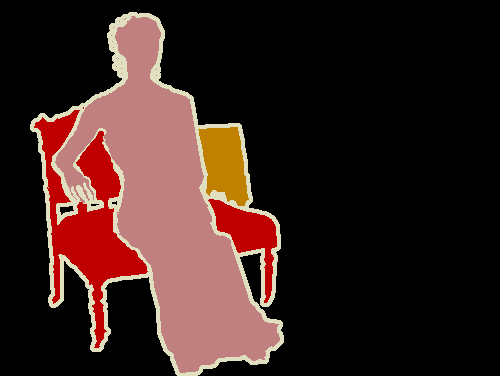} &
			\includegraphics[width=32mm,height=21mm]{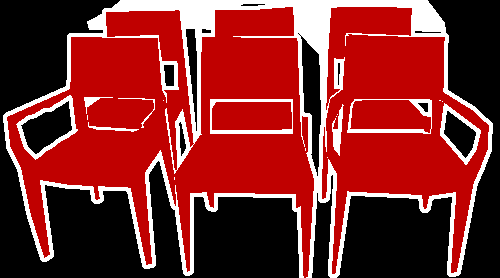} \\
			

		\end{tabular}
		\vspace{3pt}
		\caption{Segmentation results on PASCAL VOC 2012 validation set. From the first row to the last, they are (a) Input images, (b) DeepLab-LargeFOV~\cite{chen14semantic} with fully supervised manner, (c) DeepLab-LargeFOV by ours with bounding box annotations, (d) DeepLab-ResNet101~\cite{chen_deeplabv2} with fully supervised manner, (e) DeepLab-ResNet101 by ours with bounding box annotations, (f) Ground-truth. The results by Ours are even comparable with those by fully supervised segmentation models. }
		\label{fig:results}
		\vspace{-1.0em}
	\end{figure*}

	\begin{table}[!t]
		\centering
		\caption{Comparison of segmentation results by adopting DeepLab-LargeFOV~\cite{chen14semantic} as segmentation backbone on VOC 2012 validation set.
			The mIoU values of competing methods are reported from their original papers.}
		\label{VGG-16}
		\footnotesize
		\begin{tabular}{c|c|c|c|c}
			\hline  
			
			\hline
			Supervision  & \#GT & \#Box & Methods & mIoU \\
			\hline  
			Weak (Class label) & - & - & TransferNet~\cite{hong2016learning} & 52.1 \\
			\hline
			\multirow{7}*{Weak} & \multirow{7}*{0} & \multirow{7}*{10,582} 
			& WSSL~\cite{Papandreou_2015_ICCV}  & 60.6 \\
			\cline{4-5}
			& & & BoxSup~\cite{BoxSup_Dai_2015}  & 62.0 \\
			\cline{4-5}
			& & & SDI~\cite{Khoreva_2017_SDI} & 65.7 \\
			\cline{4-5}
			& & & BCM~\cite{BCM_Song_2019_CVPR} &  66.8 \\
			\cline{4-5}
			& & & Ours$_1$ & 66.8 \\
			\cline{4-5}
			& & & Ours$_2$ &  68.3 \\
			\cline{4-5}
			& & & Ours & 68.4 \\
			\hline
			
			\hline
			
			\multirow{5}*{Semi} & \multirow{5}*{1,464} & \multirow{5}*{9,118} 
			& WSSL~\cite{Papandreou_2015_ICCV} & 65.1 \\
			\cline{4-5}
			& & & BoxSup~\cite{BoxSup_Dai_2015}  & 63.5 \\
			\cline{4-5}
			& & & SDI~\cite{Khoreva_2017_SDI} & 65.8 \\
			\cline{4-5}
			& & & BCM~\cite{BCM_Song_2019_CVPR} & 67.5 \\
			\cline{4-5}
			& & & Ours & 68.7 \\
			\hline
			
			\hline
			
			Full & 10,582 & 0 & LargeFOV~\cite{chen14semantic} & 69.6\\
			\hline
			
			\hline
		\end{tabular}
		\vspace{-1.0em}
	\end{table}
	
	\textbf{Evaluation of Pseudo Masks.}\quad
	The rationality of EM optimization is that (E step) LPG can gradually generate more precise pseudo labels to supervise (M step) the training of segmentation backbone. 
	To support it, we report mIoU values of stage-wise pseudo masks generated by LPG in Table~\ref{LPG-performance}. 
	On PASCAL VOC 2012, pseudo masks generated by our LPG are much more precise than dense-CRF, thereby improving segmentation results.

	\subsection{Comparison with State-of-the-arts}
	By respectively adopting DeepLab-LargeFOV \cite{chen_deeplabv2} and DeepLab-ResNet101 \cite{chen_deeplabv2} as semantic segmentation backbone, our method is compared with state-of-the-art methods in both weakly supervised and semi-supervised manners.
	The semantic segmentation backbone is trained with 3 stages of E-M algorithm, and the results by Ours$_1$, Ours$_2$ and Ours are produced by the trained segmentation networks from Stage 1, 2 and 3, respectively. 

	\subsubsection{DeepLab-LargeFOV as Segmentation Backbone}
	In weakly supervised manner, our method is compared with TransferNet~\cite{hong2016learning},  WSSL~\cite{Papandreou_2015_ICCV}, BoxSup~\cite{BoxSup_Dai_2015}, SDI~\cite{Khoreva_2017_SDI} and BCM~\cite{BCM_Song_2019_CVPR}, where TransferNet~\cite{hong2016learning} is trained by image-level class labels. 
	The quantitative results are reported in Table~\ref{VGG-16}. 
	One can see that the semantic segmentation network from the first stage, \emph{i.e.}, Ours$_1$, has been comparable with all the competing methods. 
	The reason can be attributed to LPG for generating more accurate segmentation masks for training the segmentation network, while the other competing methods adopt generic proposal generators, \emph{e.g.}, dense CRF, MCG or GrabCut. 
	Furthermore, the training of segmentation backbones in multiple stages leads to significant performance gains, due to that LPG can gradually produce much more accurate pixel-level pseudo segmentation masks as supervision.  
	One may also find that the performance gap of box-supervised segmentation is much closer to ground-truth pixel-supervised segmentation results, \emph{i.e.}, 68.4 mIoU (Weak) v.s. 69.6 mIoU (Full) in Table \ref{VGG-16}.
	Since all these weakly supervised methods did not release source codes, we show the segmentation results by Ours in comparison to those with fully supervised manner, as shown in Fig. \ref{fig:results}. 
	One can see that our method trained with only box-supervision can produce comparable segmentation results with fully supervised DeepLab-LargeFOV model.
	Moreover, TransferNet~\cite{hong2016learning} is taken into comparision. Since it is based on image-level class labels, its segmentation performance is significantly inferior to those by box-supervision.

	In semi-supervised manner, all the competing methods obtain performance gains than themselves in weakly supervised manner.
	That is to say, with only 1,464 ground-truth pixel-annotations, the semantic segmentation performance can be notably boosted. 
	We can conclude that more accurate masks would lead to better segmentation performance.
	It is noteworthy that our method in weakly supervised manner is even superior to all the other competing methods in semi-supervised manner, which supports that our LPG can generate more accurate pseudo masks. 
	%
	%
	
	\begin{table}[!t]
		\centering
		\caption{Comparison of segmentation results by adopting DeepLab-ResNet101~\cite{chen14semantic} as segmentation backbone on VOC 2012 validation set.
			The mIoU values of competing methods are reported from their original papers.}
		\label{ResNet-101}
		\small
		\begin{tabular}{c|c|c|c|c}
			\hline  
			
			\hline
			Supervision  & \#GT & \#Box & Methods & mIoU \\
			\hline 
			
			\hline  
			\multirow{5}*{Weak} & \multirow{5}*{0} & \multirow{5}*{10,582} 
			& SDI~\cite{Khoreva_2017_SDI} & 69.4 \\
			\cline{4-5}
			& & & BCM~\cite{BCM_Song_2019_CVPR} &  70.2 \\
			\cline{4-5}
			& & & Ours$_1$ & 71.6 \\
			\cline{4-5}
			& & & Ours$_2$ &  73.1 \\
			\cline{4-5}
			& & & Ours & 73.3 \\
			\hline
			
			\hline
			
			\multirow{2}*{Semi} & \multirow{2}*{1,464} & \multirow{2}*{9,118} 
			& BCM~\cite{BCM_Song_2019_CVPR} & 71.6 \\
			\cline{4-5}
			& & & Ours & 73.8 \\
			\hline
			
			\hline
			
			Full & 10,582 & 0 & ResNet101~\cite{chen_deeplabv2} & 74.5\\
			\hline
			
			\hline
		\end{tabular}
		\vspace{-0.5em}
	\end{table}
	
	\begin{table}[!t]\small
		\setlength\tabcolsep{10pt}
		\centering
		\caption{Mean IoU of segmentation results by Deeplab-LargeFOV~\cite{chen14semantic} on PASCAL VOC 2012 validation set. LPG$_{full}$ is trained on COCO (60 classes) with full training set, while $\text{LPG}_{30\%}$, $\text{LPG}_{60\%}$ and LPG$_{85\%}$ are trained on COCO (60 classes) with $30\%$, $60\%$ and $85\%$ training masks.}
		\label{fewer_images}
		\begin{tabular}{c|c|c|c|c}
			\hline
			
			\hline
			& Stage-1 & Stage-2 & Stage-3  & Stage-4 \\
			\hline
			$\text{LPG}_{30\%}$ & 64.1 & 65.6 & 65.7 & 65.7 \\
			\hline
			$\text{LPG}_{60\%}$ & 65.7 & 67.3 & 67.5 & 67.6 \\
			\hline
			$\text{LPG}_{85\%}$ & 66.3 & 67.8 & 68.0 & 67.9   \\
			\hline
			$\text{LPG}_{full}$ & 66.8 & 68.3 & 68.4 & 68.3   \\
			\hline
			
			\hline
			
			\hline
		\end{tabular}
		\vspace{-0.5em}
	\end{table}
	
	\begin{table}[!t]\small
		\vspace{-0.6em}
		\setlength\tabcolsep{14pt}
		\centering
		\caption{Mean IoU of segmentation results by LargeFOV on CO- CO validation set (60 classes). LPG$_{coco}$ is trained on COCO (60 classes), while LPG$_{voc}$ is trained on PASCAL VOC (20 classes).}
		\label{VOC-LPG}
		\begin{tabular}{c|c|c|c}
			\hline
			
			\hline
			& Stage-1 & Stage-2 & Stage-3  \\
			\hline
			$\text{LPG}_{coco}$ & 42.1 & 43.2 & 43.4   \\
			\hline
			$\text{LPG}_{voc}$ & 40.3 & 40.9 & 40.8 \\
			\hline
			
			\hline
		\end{tabular}
		\vspace{-1.0em}
	\end{table}
	
	\begin{table}[hbt]
		\centering
		\caption{Comparison of segmentation results by adopting DeepLab-ResNet101~\cite{chen14semantic} as segmentation backbone on VOC 2012 validation set with different LPG settings. $\text{LPG}^{\prime}$: Trained on COCO-60 specified for DeepLab-LargeFOV~\cite{chen14semantic}, deployed to VOC 2012 for training DeepLab-ResNet101~\cite{chen_deeplabv2}. LPG: Trained on COCO-60 specified for DeepLab-ResNet101~\cite{chen_deeplabv2}, deployed to VOC 2012 for training DeepLab-ResNet101~\cite{chen_deeplabv2}.}
		\label{LPG-settings}
		
		\small
		\begin{tabular}{p{60pt}<{\centering}|p{15pt}<{\centering}|p{32pt}<{\centering}|p{32pt}<{\centering}|p{32pt}<{\centering}}
			\hline  
			
			\hline
			LPG-Settings  & \#GT & \#Box & Stages & mIoU \\
			\hline  
			
			\hline
			\multirow{3}*{$\text{LPG}^{\prime}$} & \multirow{3}*{0} & \multirow{3}*{10,582} &
			Ours${^\prime}_1$ & 71.3 \\
			\cline{4-5}
			& & &  Ours${^\prime}_2$ &  72.6 \\
			\cline{4-5}
			& & &  Ours$^\prime$ & 72.9 \\
			\hline
			
			\hline
			
			\multirow{3}*{LPG} & \multirow{3}*{0} & \multirow{3}*{10,582} &
			Ours$_1$ & 71.6 \\
			\cline{4-5}
			& & &  Ours$_2$ &  73.1 \\
			\cline{4-5}
			& & & Ours & 73.3 \\
			\hline
			
			\hline
			
			\hline
		\end{tabular}
		\vspace{-0.5em}
	\end{table}

	\begin{figure*}[!hbt]
		\small
		\setlength{\tabcolsep}{2.0pt}
		\centering
		\begin{tabular}{cccccc}
			\includegraphics[width=28mm,height=20mm]{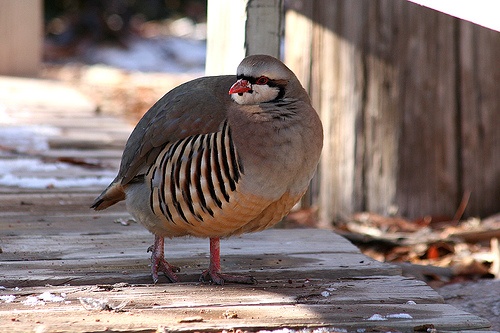} &
			\includegraphics[width=28mm,height=20mm]{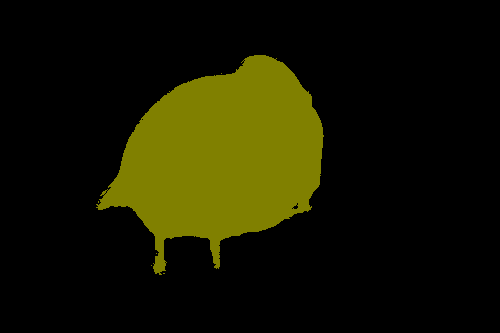} &
			\includegraphics[width=28mm,height=20mm]{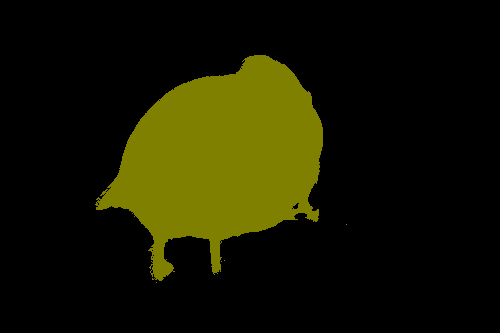} &
			\includegraphics[width=28mm,height=20mm]{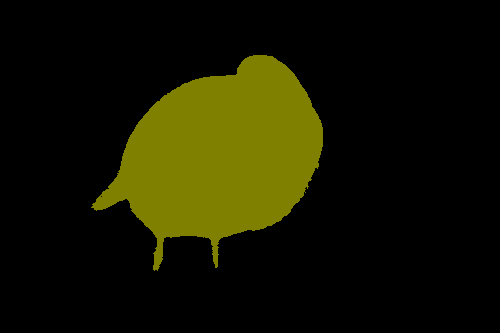} &
			\includegraphics[width=28mm,height=20mm]{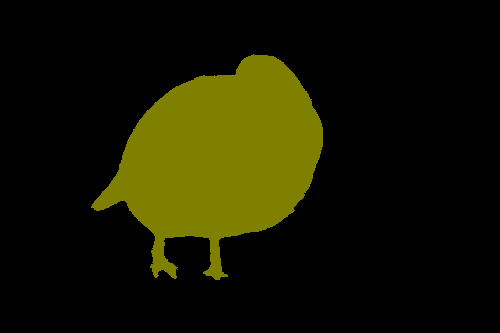} &
			\includegraphics[width=28mm,height=20mm]{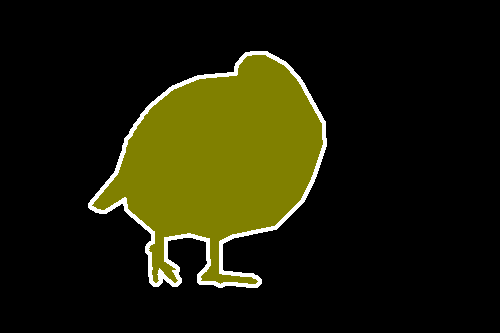}  \\
			
			\includegraphics[width=28mm,height=20mm]{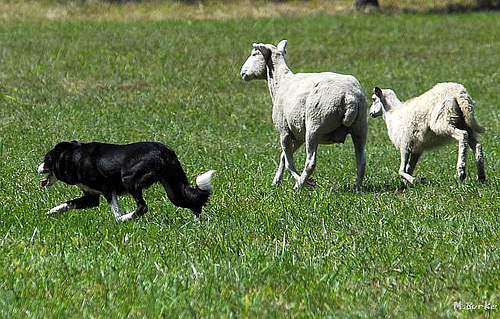} &
			\includegraphics[width=28mm,height=20mm]{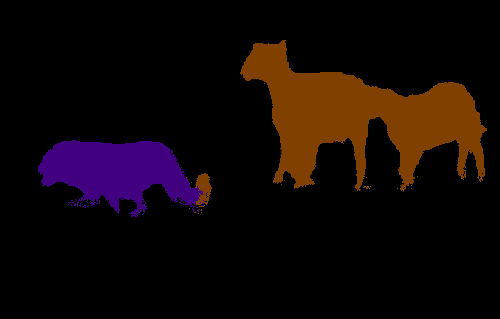} &
			\includegraphics[width=28mm,height=20mm]{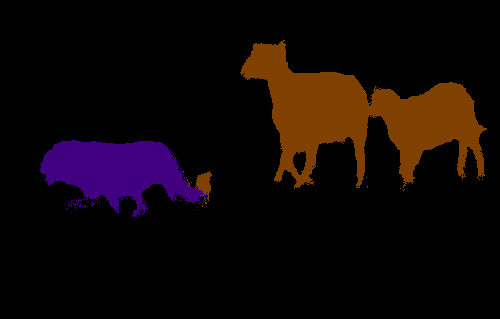} &
			\includegraphics[width=28mm,height=20mm]{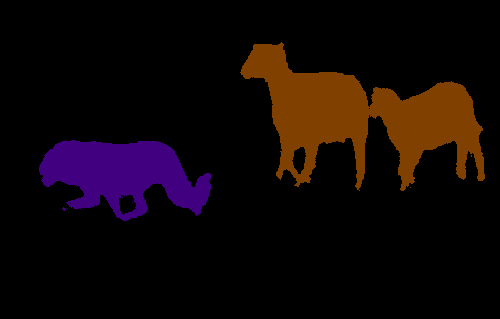} &
			\includegraphics[width=28mm,height=20mm]{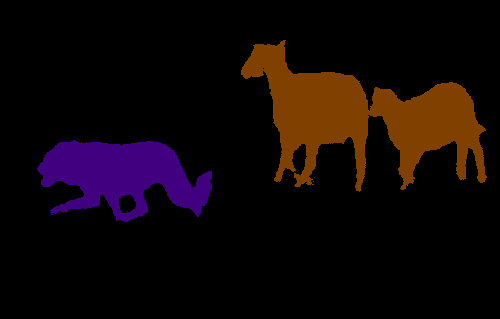} &
			\includegraphics[width=28mm,height=20mm]{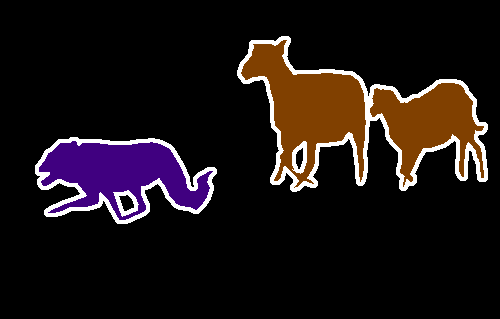}  \\
			
			\includegraphics[width=28mm,height=20mm]{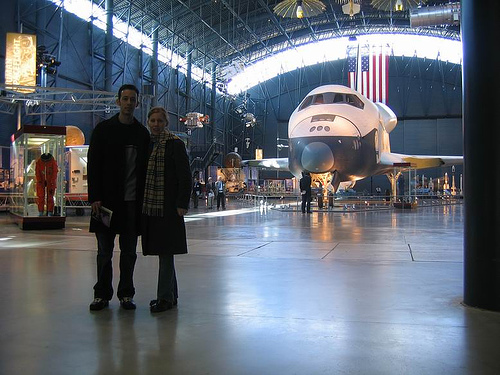} &
			\includegraphics[width=28mm,height=20mm]{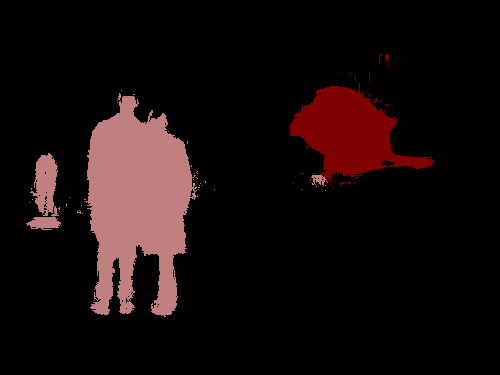} &
			\includegraphics[width=28mm,height=20mm]{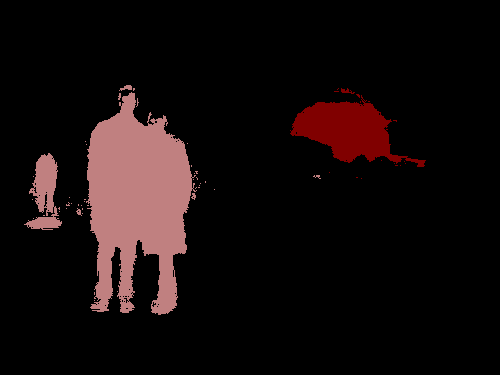} &
			\includegraphics[width=28mm,height=20mm]{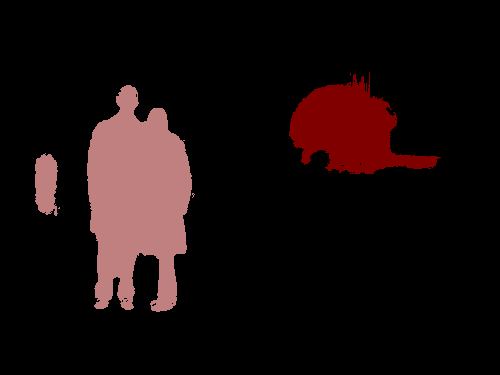} &
			\includegraphics[width=28mm,height=20mm]{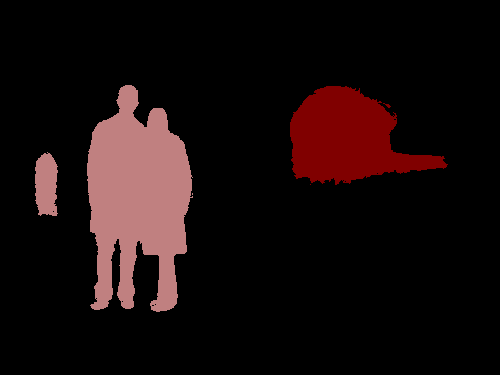} &
			\includegraphics[width=28mm,height=20mm]{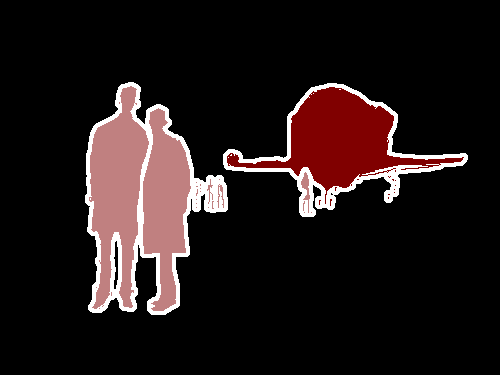}  \\
			
			\includegraphics[width=28mm,height=20mm]{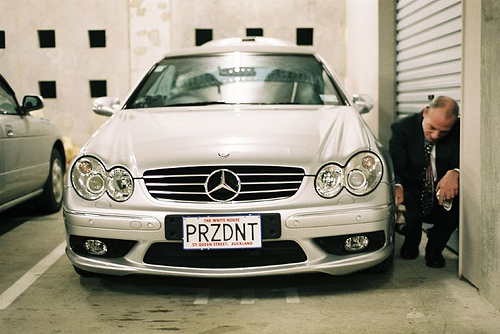} &
			\includegraphics[width=28mm,height=20mm]{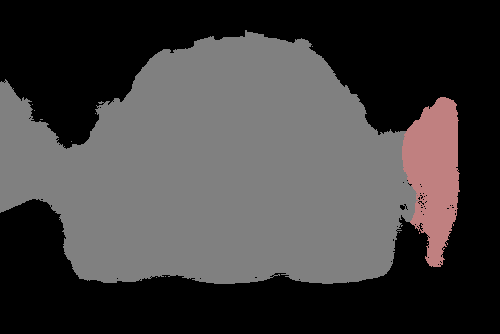} &
			\includegraphics[width=28mm,height=20mm]{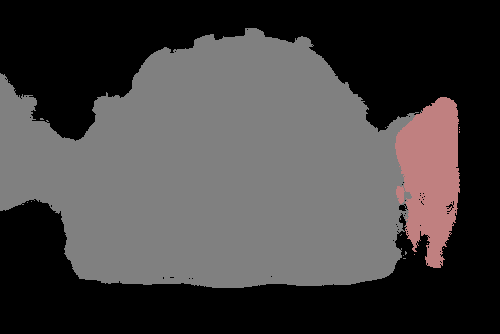} &
			\includegraphics[width=28mm,height=20mm]{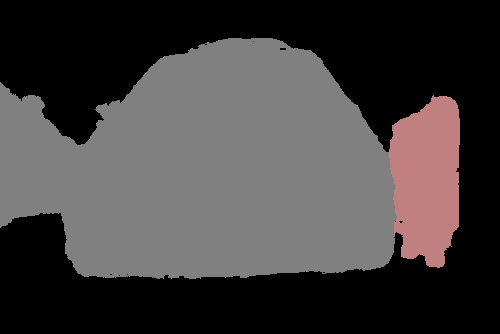} &
			\includegraphics[width=28mm,height=20mm]{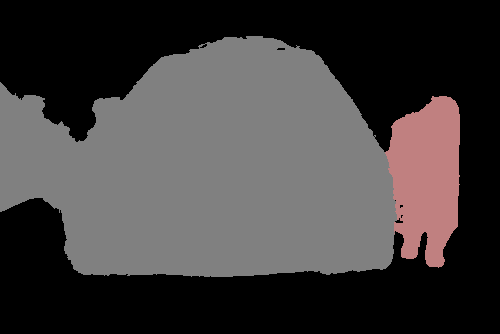} &
			\includegraphics[width=28mm,height=20mm]{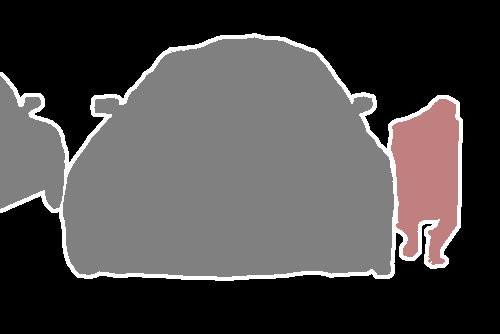}  \\
			
			\includegraphics[width=28mm,height=20mm]{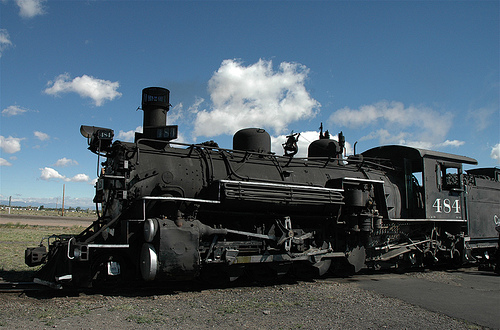} &
			\includegraphics[width=28mm,height=20mm]{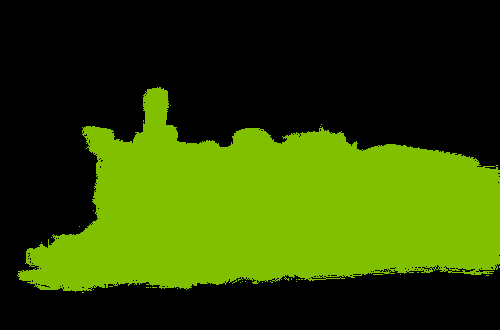} &
			\includegraphics[width=28mm,height=20mm]{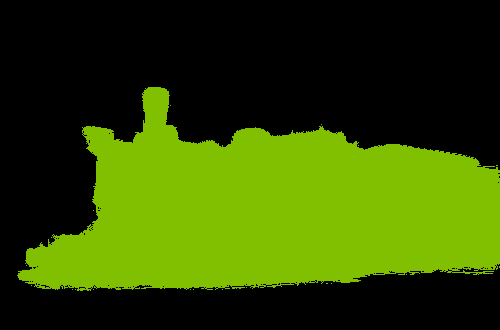} &
			\includegraphics[width=28mm,height=20mm]{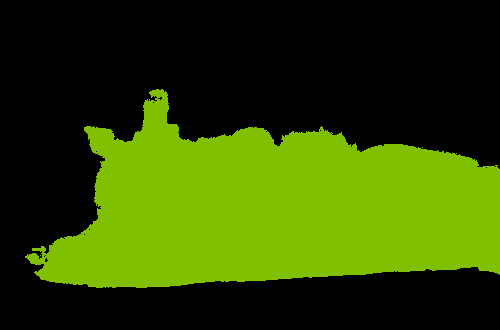} &
			\includegraphics[width=28mm,height=20mm]{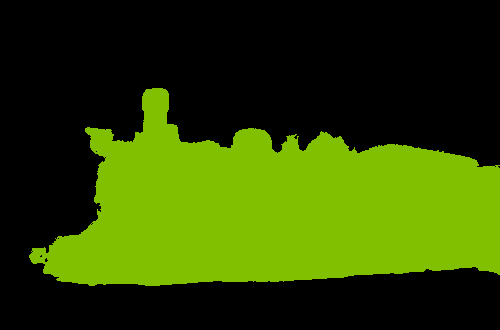} &
			\includegraphics[width=28mm,height=20mm]{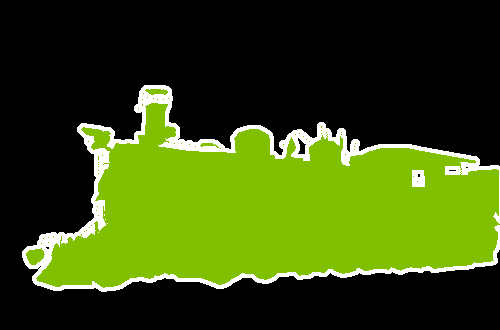}  \\
			
			\includegraphics[width=28mm,height=20mm]{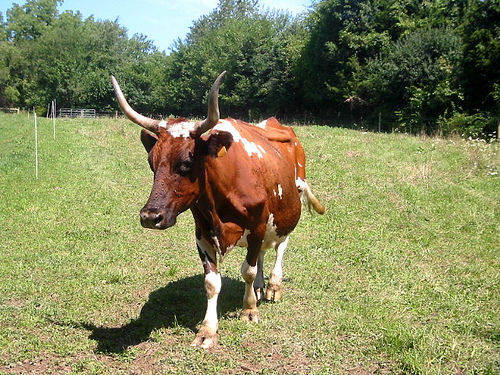} &
			\includegraphics[width=28mm,height=20mm]{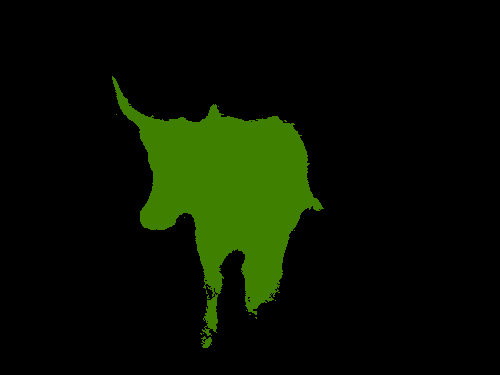} &
			\includegraphics[width=28mm,height=20mm]{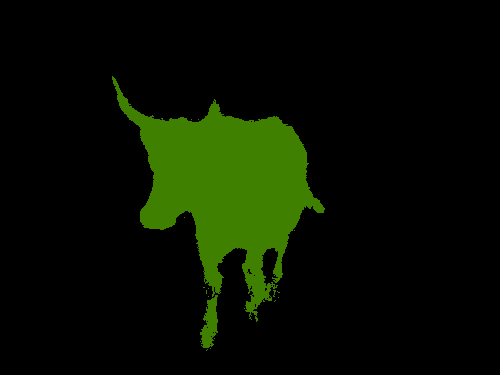} &
			\includegraphics[width=28mm,height=20mm]{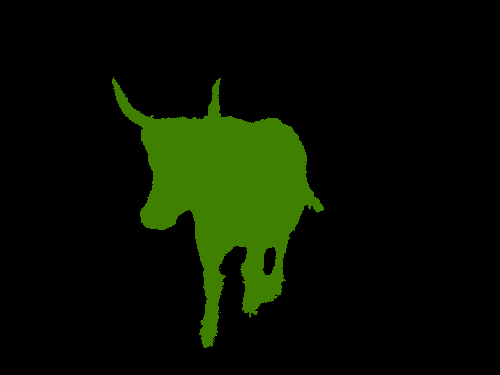} &
			\includegraphics[width=28mm,height=20mm]{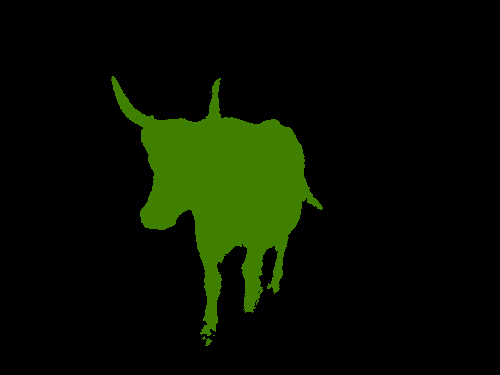} &
			\includegraphics[width=28mm,height=20mm]{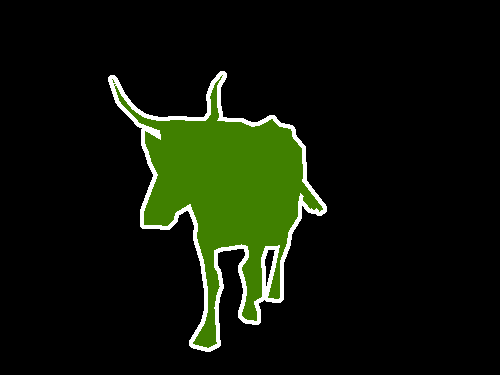}  \\
			
			\includegraphics[width=28mm,height=40mm]{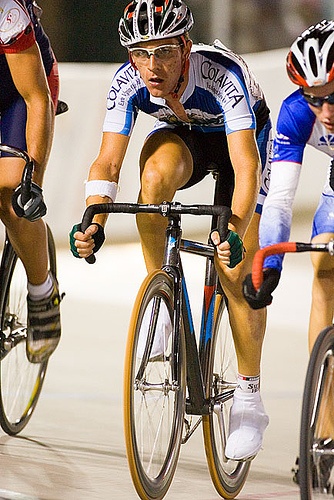} &
			\includegraphics[width=28mm,height=40mm]{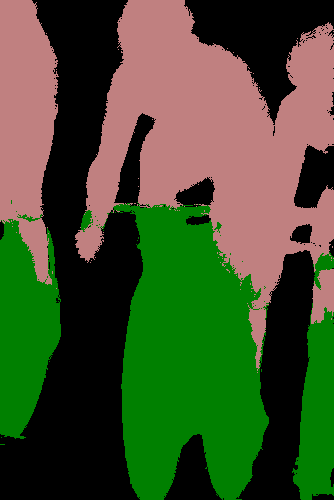} &
			\includegraphics[width=28mm,height=40mm]{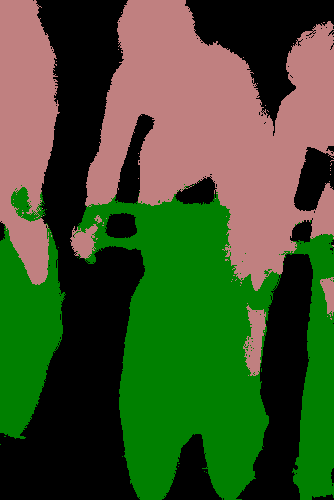} &
			\includegraphics[width=28mm,height=40mm]{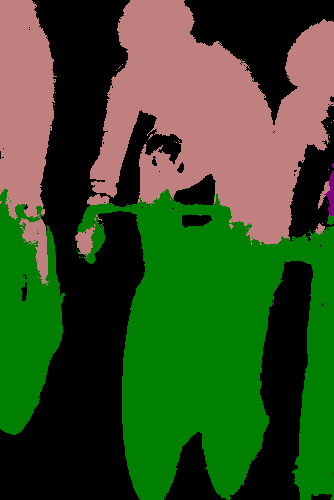} &
			\includegraphics[width=28mm,height=40mm]{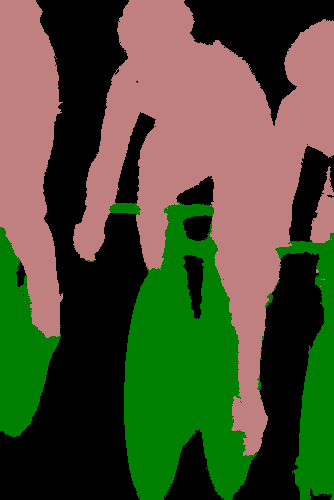} &
			\includegraphics[width=28mm,height=40mm]{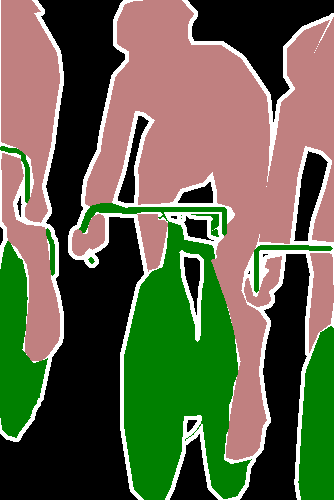}  \\
			
			\includegraphics[width=28mm,height=40mm]{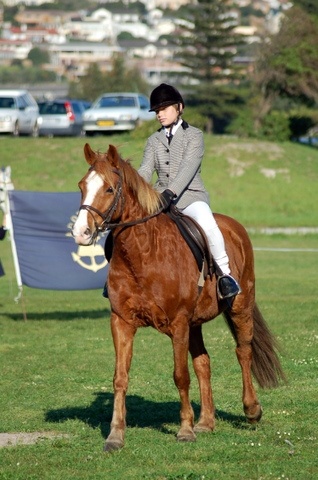} &
			\includegraphics[width=28mm,height=40mm]{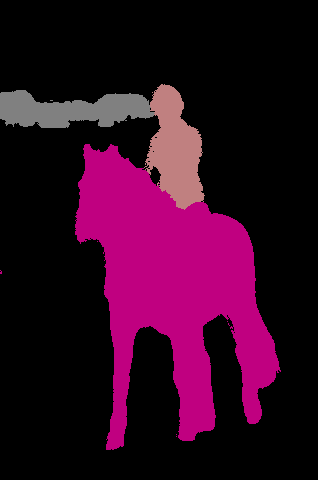} &
			\includegraphics[width=28mm,height=40mm]{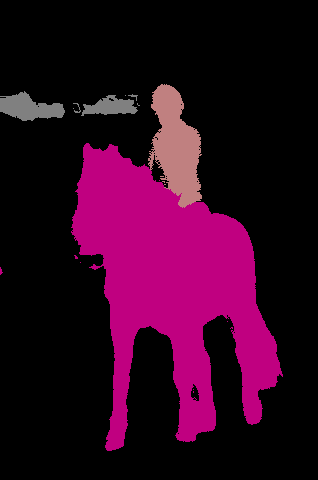} &
			\includegraphics[width=28mm,height=40mm]{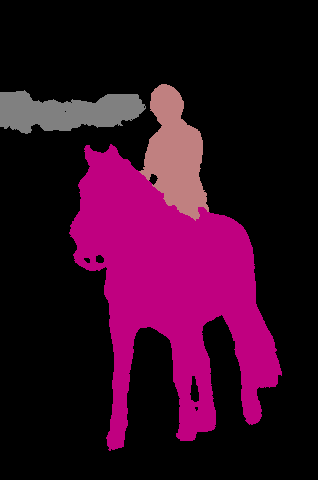} &
			\includegraphics[width=28mm,height=40mm]{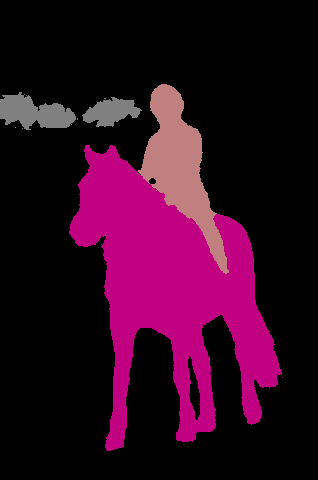} &
			\includegraphics[width=28mm,height=40mm]{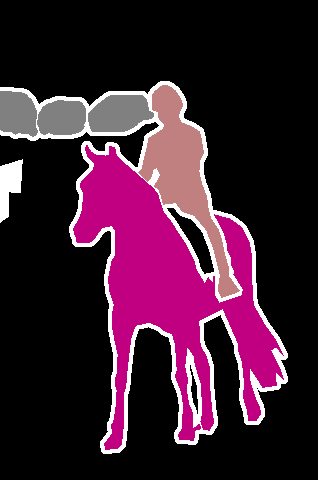}  \\

			Input images  & LargeFOV~\cite{chen14semantic} (Ours) & LargeFOV~\cite{chen14semantic} (Fully) & ResNet101~\cite{chen_deeplabv2} (Ours) & ResNet101~\cite{chen_deeplabv2} (Fully) & Ground-truth \\

		\end{tabular}
		\caption{Segmentation results in comparison to fully supervised models on PASCAL VOC 2012 validation set. The results by Ours are comparable with those by fully supervised segmentation models.}
		\label{fig:results_suppl}
		\vspace{-0.5em}
	\end{figure*}
	
	\subsubsection{DeepLab-ResNet101 as Segmentation Backbone}
	In weakly supervised manner, our method is compared with SDI~\cite{Khoreva_2017_SDI} and BCM~\cite{BCM_Song_2019_CVPR}.  
	The comparison results are reported in Table \ref{ResNet-101}. 
	Due to stronger modeling capacity of ResNet101~\cite{resnet_he}, the segmentation performance has been improved than that by DeepLab-LargeFOV~\cite{chen14semantic}. 
	Our segmentation network from the first stage has achieved $1.3\%$ mIoU gain over state-of-the-art BCM, and the gain is further enlarged to $2.6\%$ mIoU by our segmentation network from Stage 3. 
	Fig. \ref{fig:results} shows the segmentation results in comparison with the model trained by fully supervised manner. 
	Also the segmentation results by our method is comparable with those from fully supervised segmentation model. 
	%
	One may notice that Box2Seg \cite{box2seg} reports higher quantitative metrics than these competing methods, but it is not included into our comparison. 
	This is because Box2Seg adopts stronger segmentation backbone, and its source code and trained models are not publicly available, making it infeasible to make a fair comparison. 
	
	In semi-supervised manner, our method is only compared with BCM, and performs much better than BCM in terms of mIoU. 
	Moreover, one can see that our method in weakly supervised manner achieves mIoU gain of $1.2\%$ than BCM with semi-supervised manner, indicating the effectiveness of generating accurate pseudo masks by our LPG. 
	The visualized segmentation results in Fig.~\ref{fig:results} and Fig~\ref{fig:results_suppl} indicate that our methods are very competitive with fully supervised manner. 
	
	\begin{figure}[hbt]\small
		\setlength{\tabcolsep}{2.0pt}
		\centering
		\begin{tabular}{cccc}
			\includegraphics[width=.23\linewidth]{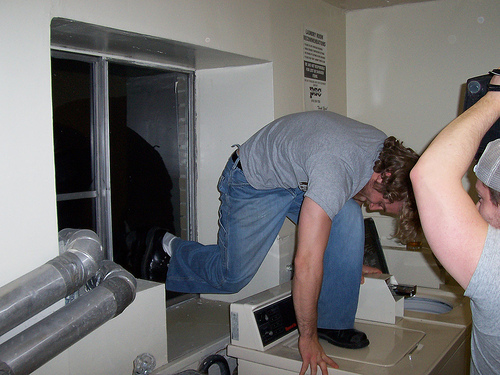}  &
			\includegraphics[width=.23\linewidth]{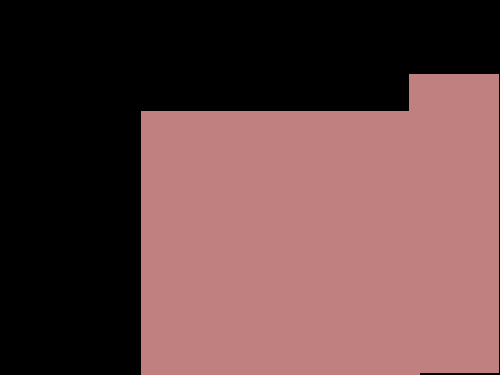}  &
			\includegraphics[width=.23\linewidth]{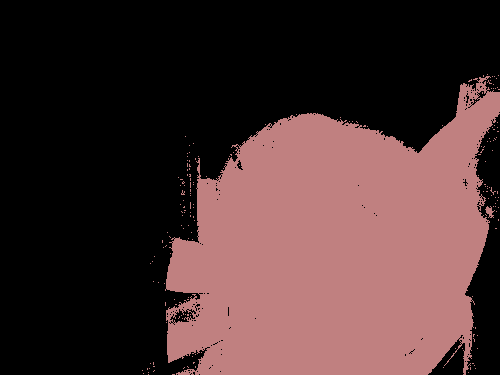}  &
			\includegraphics[width=.23\linewidth]{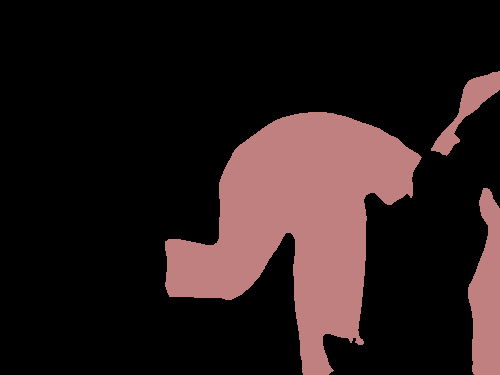}  \\
			
			\includegraphics[width=.23\linewidth]{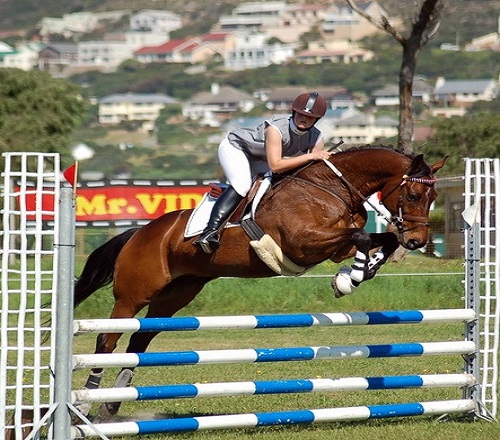}  &
			\includegraphics[width=.23\linewidth]{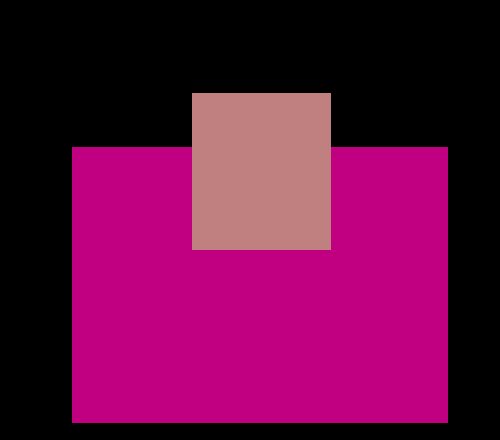}  &
			\includegraphics[width=.23\linewidth]{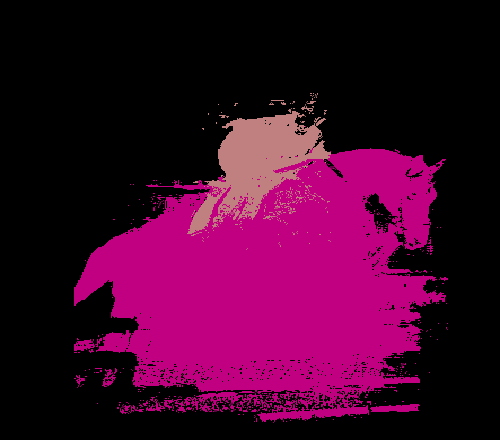}  &
			\includegraphics[width=.23\linewidth]{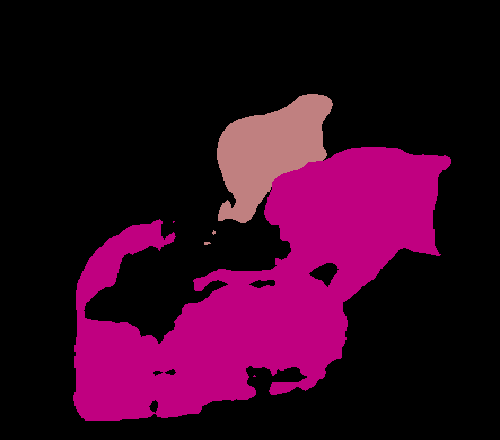}  \\

			\includegraphics[width=.23\linewidth]{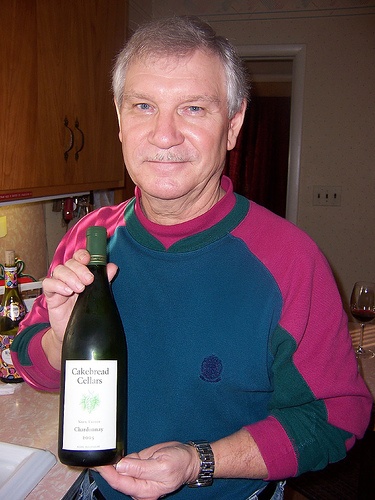}  &
			\includegraphics[width=.23\linewidth]{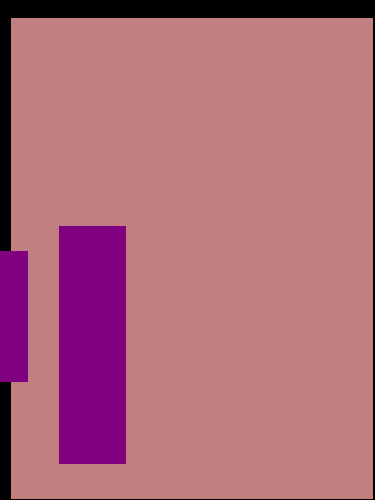}  &
			\includegraphics[width=.23\linewidth]{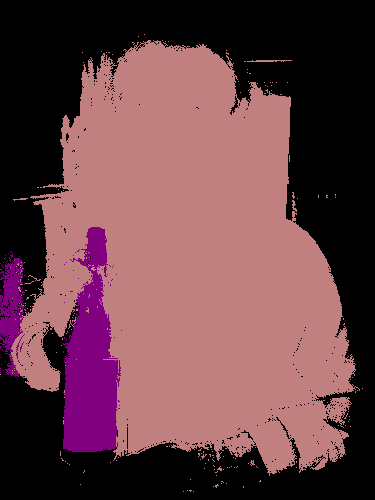}  &
			\includegraphics[width=.23\linewidth]{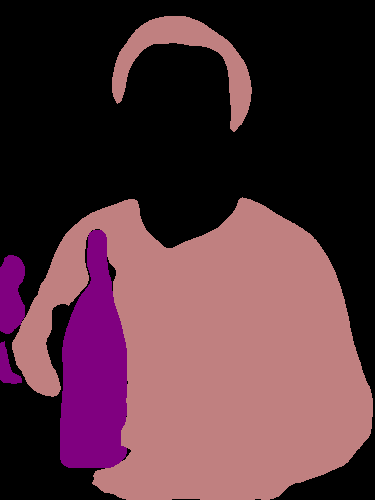}  \\
			Image & BBox &CRF~\cite{DenseCRF} & LPG \\ 
			
		\end{tabular}
		\caption{Failure cases of our LPG.}
		\vspace{-1.5em}
		\label{fig:failure-cases}
	\end{figure}
	
	\subsection{Robustness and Generalization Ability}\label{section2}
	To validate the robustness of our LPG with fewer training data, we have randomly selected $75\%$, $50\%$ and $25\%$ images from COCO (60 classes) dataset to act as $\mathcal{D}_{aux}$, resulting in $85\%$, $60\%$ and $30\%$ of segmentation masks for training.
	Table~\ref{fewer_images} shows the comparison of Deeplab-LargeFOV~\cite{chen14semantic} when collaborating with these LPG models on PASCAL VOC 2012 dataset.
	One can see that the segmentation performance will decrease with fewer training samples. 
	However, even for the least $30\%$ training samples, $\text{LPG}_{30\%}$ is inferior to $\text{LPG}_{full}$ with only about $2.7\%$ mIoU decrease, validating the robustness of our LPG.
	Moreover, existing public pixel-level annotated datasets, \emph{e.g.}, PASCAL VOC~\cite{pascal_voc}, MS-COCO~\cite{MSCOCO}, LVIS~\cite{lvis}, Open Images~\cite{openimages} \emph{etc}., are readily available to train good LPG models, offering a new feasible perspective for improving learning with weak supervision in practical applications.
	
	Furthermore, to prove the generalization ability of our LPG, we trained LPG$_{voc}$ on VOC training set (20 classes), and then applied it to COCO validation set (60 classes).
	From Table \ref{VOC-LPG}, the segmentation results by LargeFOV with LPG$_{voc}$ is only moderately inferior ($\sim$ $2\%$ by mIoU) to that with LPG$_{coco}$. 
	We note that LPG$_{coco}$ is specifically trained on COCO training set (60 classes).
	The diversity and quantity of classes, scenes and images in COCO are much greater than VOC, indicating the robustness and generalization ability of LPG.

	By ensuring that there is no intersected classes between $\mathcal{D}$ and $\mathcal{D}_{aux}$, we have validated that the trained LPG can be well generalized to new classes and new datasets.
	We further discuss whether the trained LPG models can be generalized to other segmentation backbones. 
	To this end, we train another DeepLab-ResNet101 \cite{chen_deeplabv2} segmentation network on PASCAL VOC 2012 dataset, where 3-stage LPG models (LPG$^\prime$) collaborated with DeepLab-LargeFOV \cite{chen14semantic} on MS COCO are adopted to generate pseudo masks.   
	As shown in Table~\ref{LPG-settings}, Ours$^{\prime}_1$ is inferior to Ours$_1$, since the pseudo masks generated by LPG$^\prime$ from Stage 1 are not as precise as those by LPG from Stage 1. 
	But with increasing stages, LPG$^\prime$ can generate comparable segmentation mask, and thus the final segmentation performance is very close to Ours. 
	The results show that the learned pseudo mask generators can also be well generalized to other segmentation backbones.

	\subsection{Failure Cases}
	As shown in Fig. \ref{fig:failure-cases}, LPG fails to distinguish the black shoes (left person) and the arms (right person) from background in the first image. Also some patchses near the horse (second image) are assigned as foreground, we believe the reason is due to their similar color and texture with their surroundings, and their texture discrepancy to the main body, in which case dense-CRF~\cite{DenseCRF} works even worse.

	\section{Conclusion}\label{sec:conclusion}
	
	In this paper, we proposed a learning-based pseudo mask generator (LPG) for weakly supervised semantic segmentation with bounding box annotations.
	We formulate learning pseudo mask generator as a bi-level optimization problem, and propose an EM-like algorithm to learn stage-wise and class-agnostic LPG on the auxiliary MS-COCO dataset with pixel-level and bounding box annotations. 
	Due to that LPG is specifically trained for box-supervised segmentation task, it can generate more accurate pixel-level pseudo masks, boosting the segmentation performance. 
	Experimental results on have validated the effectiveness and generalization ability of LPG.
	Our methods significantly boost the weakly supervised segmentation performance, and further close the gap between box and fully supervised learning in semantic segmentation.

	\ifCLASSOPTIONcaptionsoff
	\newpage
	\fi
	
	\bibliographystyle{IEEEtran}
	\bibliography{mybib}
	
\end{document}